\documentclass{article}
\usepackage{ijcai22}

\usepackage{times}
\usepackage{soul}
\usepackage{url}
\usepackage[hidelinks]{hyperref}
\usepackage[utf8]{inputenc}
\usepackage[small]{caption}
\usepackage{graphicx}
\usepackage{amsmath}
\usepackage{amsthm}
\usepackage{booktabs}
\usepackage{algorithm}
\usepackage{algorithmic}
\usepackage{nicefrac}  
\usepackage{multirow}
\usepackage{amssymb}
\usepackage{pifont}
\newcommand{\cmark}{\ding{51}}%
\newcommand{\xmark}{\ding{55}}%
\usepackage{tablefootnote}
\usepackage[flushleft]{threeparttable}
\usepackage{subcaption}
\usepackage{mathrsfs}
\usepackage{makecell}
\usepackage{etoolbox}
\usepackage{stfloats}
\usepackage{mathtools}
\usepackage{array}
\usepackage{lscape}
\usepackage{xcolor}
\usepackage{dsfont}

\DeclareMathOperator*{\argmin}{arg\,min}

\newcommand{\ie}{{i}.{e}.{,}}
\newcommand{\eg}{{e}.{g}.{,}}

\title{Efficient Neural Neighborhood Search for Pickup and Delivery Problems}

\author{
Yining Ma$^{1,}$\thanks{Equally contributed.}
\and
Jingwen Li$^{1, \ast}$\and
Zhiguang Cao$^{2,}$\thanks{Zhiguang Cao is the corresponding author.}\and
Wen Song$^3$\and
Hongliang Guo$^4$\and\\
Yuejiao Gong$^5$\And
Yeow Meng Chee$^1$
\affiliations
$^1$National University of Singapore\\
$^2$Singapore Institute of Manufacturing Technology, A*STAR\\
$^3$Institute of Marine Science and Technology, Shandong University\\
$^4$Institute for Infocomm Research, A*STAR\\
$^5$South China University of Technology
\emails
\{yiningma, lijingwen\}@u.nus.edu, zhiguangcao@outlook.com,
wensong@email.sdu.edu.cn,
guo\_hongliang@i2r.a-star.edu.sg,
gongyuejiao@gmail.com,
ymchee@nus.edu.sg
}
\begin{document}

\maketitle

\begin{abstract}
We present an efficient Neural Neighborhood Search (N2S) approach for pickup and delivery problems (PDPs). In specific, we design a powerful Synthesis Attention that allows the vanilla self-attention to synthesize various types of features regarding a route solution. We also exploit two customized decoders that automatically learn to perform removal and reinsertion of a pickup-delivery node pair to tackle the precedence constraint. Additionally, a diversity enhancement scheme is leveraged to further ameliorate the performance. Our N2S is generic, and extensive experiments on two canonical PDP variants show that it can produce state-of-the-art results among existing neural methods. Moreover, it even outstrips the well-known LKH3 solver on the more constrained PDP variant. Our implementation for N2S is available online\footnote{Code is available at \href{https://github.com/yining043/PDP-N2S}{https://github.com/yining043/PDP-N2S}. The published version of this preprint can be found at IJCAI-22 (see \href{https://www.ijcai.org/proceedings/2022/662}{https://www.ijcai.org/proceedings/2022/662}). This preprint fixes some typos and contains additional discussions and results.
}.
\end{abstract}

\section{Introduction}
\label{sec:intro}
\begin{figure*}
     \centering
     \includegraphics[width = 0.999\textwidth]{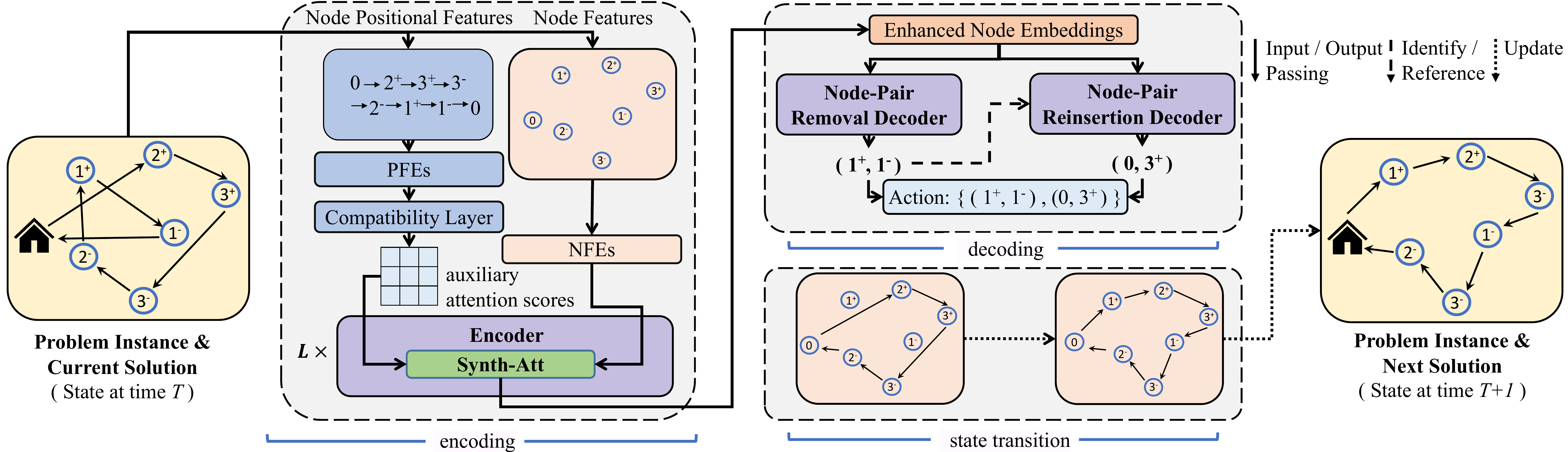}
     \caption{ An example of a PDTSP-7 instance to illustrate our N2S approach. From the left to the right: encoding, decoding and state transition process. The encoders process raw features of the current solution to produce node embeddings, which are then fed into the two decoders to sample an action. In the state transition, the node pair ($1^+,1^-$) are removed and then reinserted after depot (0) and node $3^+$, respectively.}
     \label{fig:framework}
\end{figure*}

Efficient neighborhood search functions as the key component of powerful heuristics for pickup and delivery problems (PDPs)~\cite{parragh2008survey}. It involves an iterative search process which transforms a solution into another candidate in its current neighborhood, hopefully in an efficient way. Designing the neighborhood and search rules usually determines the efficiency and even the success of a solver. 
However, they are often problem-specific and \emph{manually engineered} with a lot of trial and error, which needs to redesign when changes occur in constraints or objectives. These limitations may hinder the applications in the rapidly evolving industries.

On the other hand, recent neural methods for vehicle routing problems (VRPs) have emerged as promising alternatives to traditional heuristics (\eg~\cite{ma2021learning}). They are usually faster, and more importantly, could automate the design of heuristics for new variants where no hand-crafted rule is available~ \cite{kwon2020pomo}. However, the prevailing neural methods mainly focus on travelling salesman problem (TSP) or capacitated vehicle routing problem (CVRP), where efficient solvers for PDPs are rarely studied. The PDP is ubiquitous in logistics, robotics, meal-delivery services, etc~\cite{parragh2008survey}, which optimizes the route for pickup-delivery requests and is characterised by the precedence constraint (pickup before delivery). Despite the first attempt in~\cite{li2021heterogeneous}, which learns a construction method to build a PDP solution (route) in seconds, it leaves a considerable gap to traditional heuristics in solution quality.

To reduce the gap, we propose an efficient \textit{{N}eural {N}eighborhood {S}earch ({N2S})} approach for PDPs, based on a novel Transformer styled policy network with the encoder-decoder structure. To our knowledge, the most similar existing policy network to ours is the DACT in \cite{ma2021learning}. Also as an improvement method, DACT learns to encode the current solution and transform it into another one using the \emph{2-opt}, \emph{insert}, or \emph{swap} decoder. Among them, the \emph{2-opt} decoder, which considers reversing a segment of the solution, performed the best for TSP and CVRP. However, it is not suitable for PDPs since the precedence constraint can be easily violated by segment inversion. Though the \emph{insert} decoder performed well on small-scale PDPs, which considers removing and reinserting a single node, its performance significantly drops on larger-scale or highly constrained PDPs (see Section~\ref{sec:exp}). Conversely, our N2S tackles the precedence constraint more efficiently by allowing a pair of pickup-delivery nodes to be simultaneously operated in the neighborhood search through two customized \emph{removal} and \emph{reinsertion} decoders as shown in Figure~\ref{fig:framework}.

Another challenge lies in the design of the encoders. It was well revealed in \cite{ma2021learning} that the vanilla Transformer encoder~\cite{vaswani2017attention} failed to correctly encode route solutions since the embeddings of node features (\ie~coordinates) and node positional features (\ie~node positions) involve two different \emph{aspects} of a route solution which are not directly compatible during encoding. They thus proposed the dual-aspect collaborative attention (DAC-Att) to learn dual representations for each feature \emph{aspect}. In this paper, we propose a simple yet powerful \emph{{Synth}esis {Att}ention ({Synth-Att})} where the attention scores from various types of node feature embeddings can be synthesized to attain a comprehensive representation. It not only has the potential to encode more \emph{aspects} than DAC-Att, but also reduces the computation costs while reserving competitive performance.

Additionally, we design a diversity enhancement scheme to further ameliorate the performance. The proposed N2S approach was trained through reinforcement learning~\cite{ma2021learning}, and we evaluated it on two canonical problems in the PDP family, \ie~the pickup and delivery travelling salesman problem (PDTSP) and its variant with the last-in-first-out constraint (PDTSP-LIFO) to verify our design. Experimental results show that our N2S outperforms the state-of-the-art, and becomes the \emph{first} neural method to surpass the well-known LKH3 solver~\cite{lkh3} when solving the synthesized PDP instances.

The contributions of the paper are summarized as follows: 1) we present an efficient N2S approach for PDPs, which learns to perform removal and reinsertion of a pickup-delivery node pair automatically; 2) we propose Synth-Att, which allows the vanilla self-attention mechanism to synthesize node relationships from various feature embeddings in a simple way, and achieves superior expressiveness with much fewer computation costs in comparison to DAC-Att; 3) we explore a diversity enhancement scheme, which further makes our N2S the \emph{first} neural method with almost no domain knowledge to surpass the LKH3 solver on the synthesized PDP instances (when there are no significant distribution shift w.r.t training ones). All these advantages highlight the potential of N2S as a powerful neural solver for new PDP variants without much need for manual trial and error.

\section{Related Work}
\label{sec:relatedwork}

\paragraph{Neural Methods for VRPs.}
We classify recent \emph{neural} methods into \emph{construction} and \emph{improvement} ones.
The \emph{construction} methods, e.g., \cite{nazari2018reinforcement,joshi2019efficient,kool2018attention}, learn a distribution of selecting nodes to autoregressively build solutions from scratch. Despite being fast, they lack abilities to search (near-)optimal solutions, even if armed with sampling (e.g., \cite{kool2018attention}), local search (e.g., \cite{kim2021learning}), or Monte-Carlo tree search (e.g., \cite{fu2020generalize}). Among them, POMO \cite{kwon2020pomo} which explored diverse rollouts and data augments is recognized as the best construction method. Differently, \emph{improvement} methods often hinge on a neighborhood search procedure such as \emph{node swap} in \cite{chen2019learning}, \emph{ruin-and-repair} in \cite{hottung2019neural}, and \emph{2-opt} in \cite{wu2021learning}. \cite{ma2021learning} extended the Transformer styled model of \cite{wu2021learning} to Dual-Aspect Collaborative Transformer (DACT), and achieved the state-of-the-art performance, which was also competitive to the \emph{hybrid} neural methods, e.g., the ones combined with differential evolution \cite{hottunglearning} and dynamic programming \cite{kool2021deep}. Despite the success of the above methods for CVRP or TSP, they are not verified on the precedence constrained PDPs. Though \cite{li2021heterogeneous} made the first attempt to learn a construction solver for PDPs by introducing the heterogeneous attention to \cite{kool2018attention}, the resulting solution qualities are still far from the optimality.

\paragraph{Neighbourhood Search for PDPs.} Various heuristics based on neighborhood search have been proposed for PDPs. For PDTSP,
\cite{savelsbergh1990efficient} studied the \emph{$k$-interchange} neighborhood. A \emph{ruin-and-repair} neighborhood was later proposed in \cite{renaud2000heuristic}, and further extended in \cite{renaud2002perturbation} with multiple perturbation methods. 
Aside from PDTSP, other PDP variants were also studied, normally solved by designing new problem-specific neighborhoods \cite{parragh2008survey}, e.g., \cite{veenstra2017pickup} proposed five neighborhoods to tackle the PDTSP with handling costs. Among them, the PDTSP-LIFO attracts much attention due to its largely constrained search space. To solve it, \cite{carrabs2007variable} introduced additional neighborhoods such as \emph{double-bridge} and \emph{shake}. In \cite{li2011tree}, neighborhoods with tree structures were further proposed. Different from the above PDP solvers, the well-known LKH3 solver \cite{lkh3} combines neighborhood restriction strategies to achieve more effective local search, which could solve various VRPs with superior performance. Recently, LKH3 was extended to tackle several PDP variants including PDTSP and PDTSP-LIFO, and delivered comparable performance to \cite{renaud2002perturbation} and \cite{li2011tree} on PDTSP and PDTSP-LIFO, respectively. Also given the open-sourced nature\footnote{\href{http://webhotel4.ruc.dk/~keld/research/LKH-3/}{http://webhotel4.ruc.dk/$\sim$keld/research/LKH-3/}}, we use LKH3 as the benchmark heuristic in this paper.

\section{Problem Formulation}
\label{sec:model}
\begin{figure}
     \centering
     \subfloat[$\delta_1$ (feasible to both PDPs)]{\includegraphics[width = 0.23\textwidth]{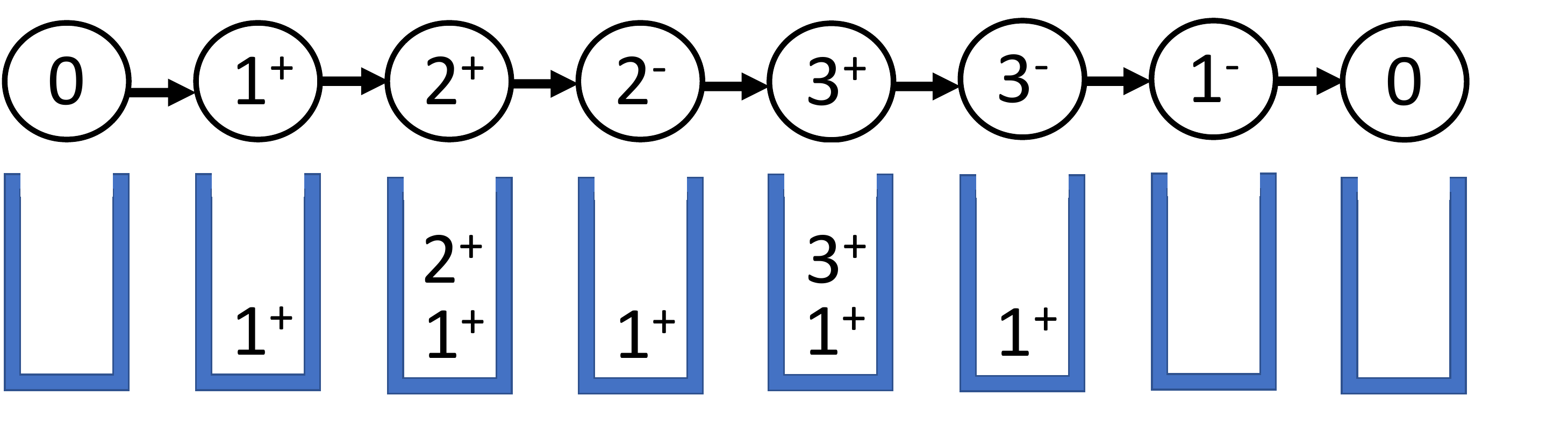}
     }
     \subfloat[$\delta_2$ (only feasible to PDTSP)]{\includegraphics[width = 0.23\textwidth]{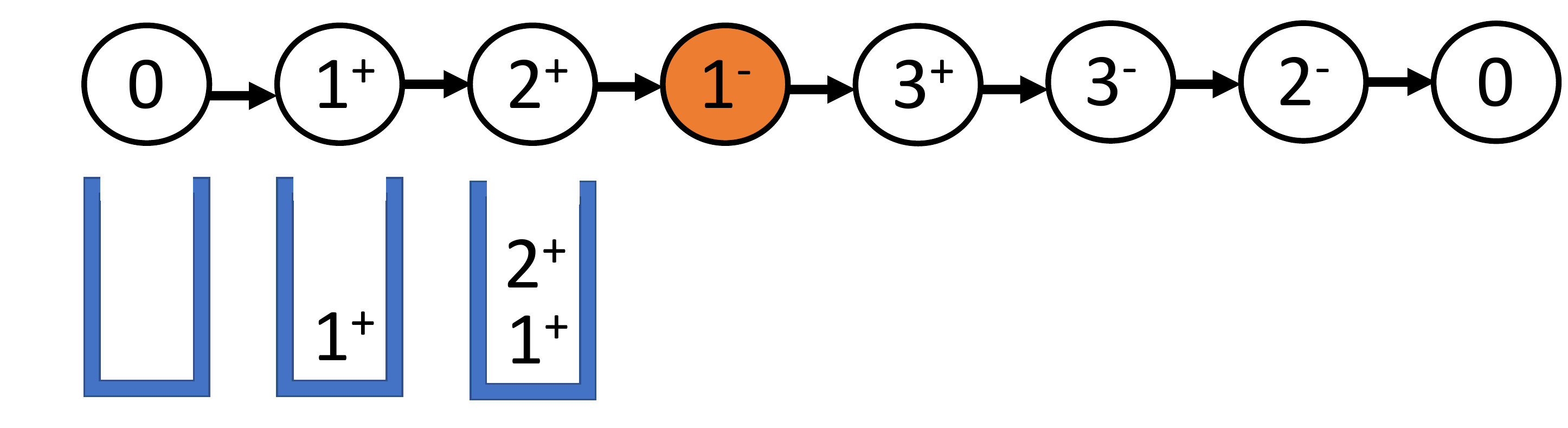}
     }
     \caption{Two PDP solutions. (a) all goods are on top of the stack at the delivery node; (b) goods from $1^+$ is blocked by $2^+$ at $1^-$.}
     \label{fig:formulation}
\end{figure}

\subsection{Notations}
We define the studied PDPs over a graph $G=(V, E)$, where nodes in $V=P\cup D\cup0$ represent locations and edges in $E$ represent trips between locations. With $n$ one-to-one pickup-delivery requests, an PDP instance contains $|V|=2n+1$ different locations, where node $0$ is depot, set $P = \{1^+,2^+,...,n^+\}$ contains pickup nodes, and set $D = \{1^-, 2^-, ..., n^-\}$ contains delivery nodes\footnote{We also refer $x_0$ to the depot, $\{x_1,...,x_n\}$ to the pickup nodes, and $\{x_{n+1},...,x_{2n}\}$ to the delivery nodes from here on.}. Each pickup node $i^+$ has a number of goods to be transported to its delivery node $i^-$. The objective is to find the shortest Hamiltonian cycle to fulfill all requests. In this paper, we consider two representative PDP variants, i.e., PDTSP and PDTSP-LIFO. The solution $\delta$ is defined as a cyclic sequence $(x_0, ..., x_{2n+1})$, where $x_0$ and $x_{2n+1}$ are the depot, and the rest is a permutation of nodes in $P\cup D$. For PDTSP, such permutation is under the \emph{precedence} constraint that requires each pickup  $i^+$ to be visited before its delivery  $i^-$. For PDTSP-LIFO, the \emph{last-in-first-out} constraint is further imposed which requires loading and unloading to be executed in the corresponding order.  This implies that unloading at a delivery node is allowed if and only if the goods is at the top of the stack. In Figure \ref{fig:formulation}, we present two example solutions with $n\!=\!3$ and $|V|\!=\!7$. The two solutions are both feasible to PDTSP, however, solution $\delta_2$ in Figure \ref{fig:formulation}(b) is infeasible to PDTSP-LIFO as the goods from $1^+$ is \emph{NOT} at the top of the stack when it needs to be delivered at $1^-$.

\subsection{MDP Formulations}
We define the process of solving PDPs by our N2S as a Markov Decision Process $\mathcal{M}\!=\!(\mathcal{S}, \mathcal{A}, \mathcal{T}, \mathcal{R}, \mathcal{\gamma})$ as follows.

\textbf{State $\mathcal{S}$}.
At time $t$, the state is defined to include: 1) features of the current solution $\delta_t$, 2) action history, and 3) objective value of the best incumbent solution, \ie 
\begin{equation}
    s_t=\left\{\{l(x)\}_{x \in V},  \{p_t(x)\}_{x \in V},  \mathcal{H}(t,K), f(\delta_t^*)\right\},
    \label{eq:states}
\end{equation}
where $\delta_t$ is described from two \emph{aspects} following \cite{ma2021learning}: $l(x)$ contains 2-dim coordinates of node $x$ (\ie~\emph{node features}) and $p_t(x)$ indicates the index position of $x$ in $\delta_t$ (\ie~\emph{node positional feature}); $\mathcal{H}(t,K)$ stores the most recent $K$ actions at time $t$ if any; and $f(\cdot)$ denotes the objective function and $\delta_t^* = \argmin_{\delta_{t'} \in \{\delta_{0}, ..., \delta_{t}\}}f(\delta_{t'})$.

\textbf{Action $\mathcal{A}$}. With action $a_t\!=\!\{ ( i^+, i^- ),  ( j, k)  \}$ where $j,k \in V \backslash \{i^+,i^-\}$, the agent removes node pair $(i^+, i^-)$, and then reinserts node $i^+$ and $i^-$ after node $j$ and $k$, respectively.

\textbf{State Transition $\mathcal{T}$}. We utilize a deterministic transition rule to perform $a_t$. An example is illustrated in Figure \ref{fig:framework}.

\textbf{Reward $\mathcal{R}$}. The reward function is defined as $r_t\!=\!f(\delta_t^{\textit{*}})\!-\!min\left[f(\delta_{t+1}), f(\delta_t^{\textit{*}})\right]$ which is the immediate reduced cost w.r.t. $f(\delta_t^{\textit{*}})$. The N2S agent aims to maximize the expected total reduced cost w.r.t. $\delta_0$ with a discount factor $\gamma < 1$.

\section{Methodology}
\label{sec:methodology}
\subsection{Encoder and Synth-Att}

Given state $s\!=\!\left\{\{l(x)\}_{x \in V}, \{p(x)\}_{x \in V}, \mathcal{H}(t,K), f(\delta^*)\right\}$, the N2S encoder takes $\{l(x)\}_{x \in V}$ and $\{p(x)\}_{x \in V}$ as inputs\footnote{$\mathcal{H}(t,K)$ is input to decoder and $f(\delta^*)$ is input to critic network (introduced later). Here we omit time step $t$ for better readability.} to learn embeddings for representing the current solution. Following DACT, we first project these raw features into two sets of embeddings, \ie~node feature embeddings (NFEs) $\{h_i\}_{i=0}^{|V|}$ and positional feature embeddings (PFEs) $\{g_i\}_{i=0}^{|V|}$. Different from \cite{ma2021learning}, we treat NFEs as the primary set of embeddings whereas PFEs as auxiliary ones.

\paragraph{NFEs.} We define $h_i$ as the linear projection of its node features $l(x_i)$ for any $x_i \in V$ with output dimension $d_{h} = 128$.

\paragraph{PFEs.} By extending the absolute positional encoding in the vanilla Transformer \cite{vaswani2017attention}, the cyclic positional encoding (CPE) was proposed in \cite{ma2021learning}, which enables Transformer to encode cyclic sequences (as our PDP solutions) more accurately. The PFE $g_i$ with output dimension $d_{g} = 128$ are initialized by CPE as follows,
\begin{equation}
\resizebox{.85\linewidth}{!}{$
  {g_i}^{(d)} = \left\{\begin{matrix}
\!sin(\omega_d\!\cdot\!\left| (z(i) \bmod \frac{4\pi}{\omega_d}) - \frac{2\pi}{\omega_d} \right|), \text{ if }d\text{ is even}\\
\!cos(\omega_d\!\cdot\!\left| (z(i) \bmod \frac{4\pi}{\omega_d}) - \frac{2\pi}{\omega_d} \right|), \text{ if }d\text{ is odd} \end{matrix}\right.
$}
\end{equation}
where superscript $d$ of ${g_i}^{(d)}$ refers to the $d$-th dimension of $g_i$, and the scalar $z(i)$ as well as the angular frequency $\omega_d = \frac{2\pi}{T_d}$ are defined accorting to Eq.~(\ref{eq:cpez}) and Eq.~(\ref{eq:cpeomiga}), respectively. 
\begin{equation}
    z(i) = \frac{i}{{|V|}}\frac{2\pi}{\omega_d}\left \lceil \frac{{|V|}}{2\pi/\omega_d} \right \rceil,
    \label{eq:cpez}
\end{equation}
\begin{equation}
\resizebox{.89\linewidth}{!}{$
T_d\!=\!\!\left\{\begin{matrix}
\frac{\!3\lfloor d/3 \rfloor + 1}{{d_g}} ({|V|}\!\!-\!\!{|V|}^{\frac{1}{\left \lfloor {d_g} / 2 \right \rfloor}})\!+\! {|V|}^{\frac{1}{\left \lfloor {d_g} / 2 \right \rfloor}}\!, \!\!\!\!&\!\text{\!if \!}d\!<\!\lfloor\!\frac{{d_g}}{2}\!\rfloor\\ 
{|V|}.\!&\!\text{otherwise}
\end{matrix}\right.
$}
\label{eq:cpeomiga}
\end{equation}

\begin{figure}
     \centering
     \subfloat[The \textbf{DAC-Att} \cite{ma2021learning}]{\includegraphics[width = 0.3\textwidth]{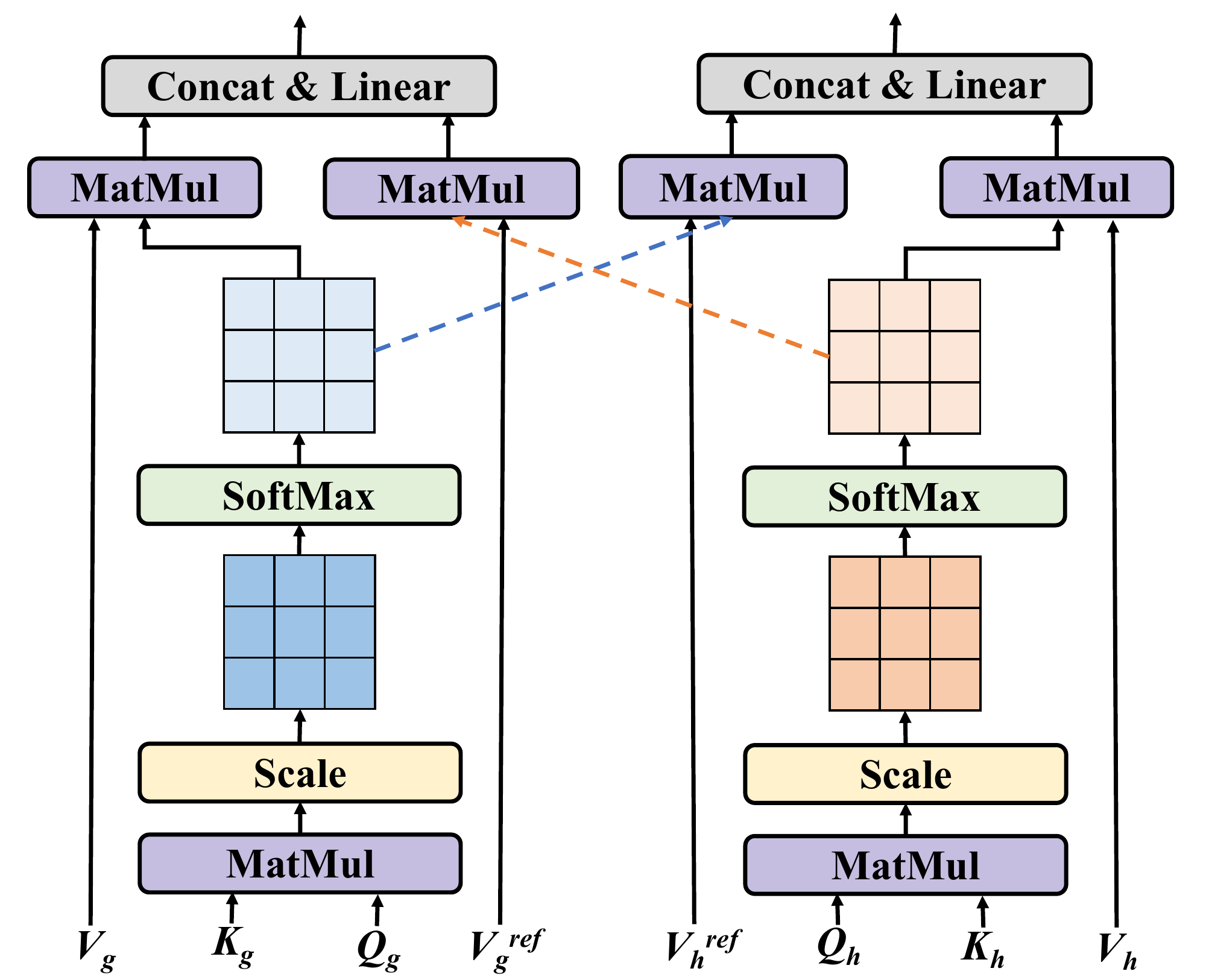}
     }
     \subfloat[Our \textbf{Synth-Att}]{\includegraphics[width = 0.175\textwidth]{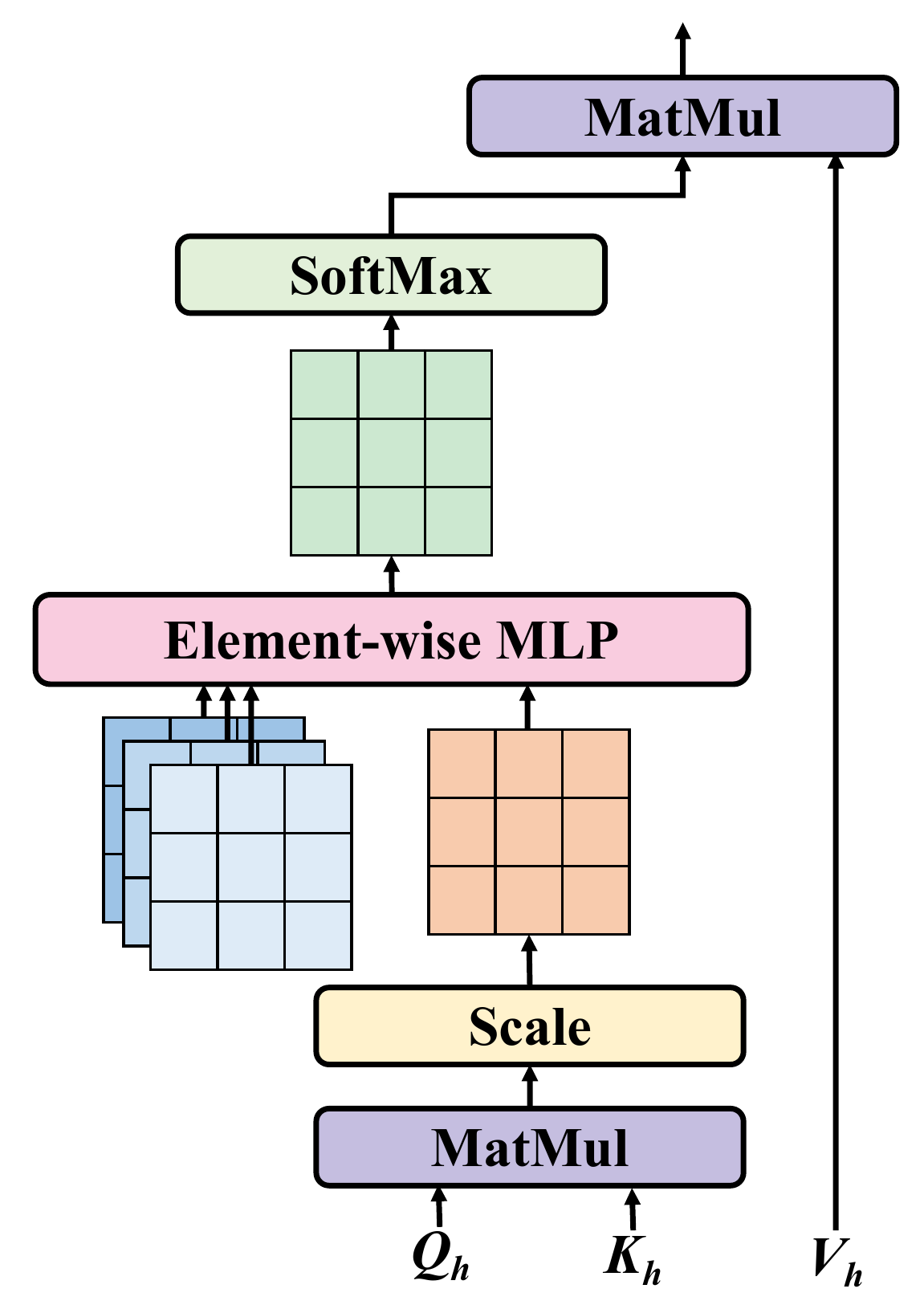}
     }
     \caption{Comparison of attention modules for VRPs. The blue and orange squares are used to represent self-attention score matrices.}
     \label{fig:attn}
\end{figure}

According to \cite{ma2021learning}, directly fusing the two sets of embeddings (\ie~$h_i\!+\!g_i$) may cause undesired noises to the vanilla self-attention. As shown in Figure \ref{fig:attn}(a), they thus proposed the DAC-Att in ways that each embedding set independently computes attention scores and shares the other \emph{aspect} to learn dual-aspect representations. Different from it, we propose a simple and generic mechanism by incorporating a multilayer perception (MLP). As shown in Figure \ref{fig:attn}(b), besides the original self-attention scores (the orange squares), multiple auxiliary attention scores learned from other feature embeddings (the blue squares) are leveraged and fed into an element-wise MLP, which allows it to synthesize heterogeneous attention relationships into comprehensive ones. We call it \emph{\textbf{Synth}esis \textbf{Att}ention (\textbf{Synth-Att})}. It is able to not only leverage more attention scores from various feature embeddings, but also achieve competitive performance to DAC-Att with less computation costs. Below, we present more details. 

\paragraph{Auxiliary Attention Scores.}
In our N2S, PFEs are used to generate multi-head auxiliary attention scores as follows,
\begin{equation}
    \alpha_{i,j,m}^{\text{aux}} = \frac{1}{\sqrt{d_k}} \left( g_i W^{Q_\text{aux}}_{m} \right ) \left( g_j W^{K_\text{aux}}_{m} \right ) ^T,
\end{equation}
where $W^{Q_\text{aux}}_{m} \in \mathbb{R}^{d_g \times d_q}, W^{K_\text{aux}}_{m} \in \mathbb{R}^{d_g \times d_k}$ are trainable matrices for each head $m$. We set $m=4$ and $d_q = d_k = d_g / m$.

\paragraph{Syn-Att.} The Syn-Att is defined as follows,
\begin{equation}
 {\tilde h}_i = \textbf{Syn-Att} (W^{Q}, W^{K}, W^{V}, W^{O}, \text{MLP}).
\end{equation}
In specific, it first computes the multi-head self-attention scores $\alpha_{i,j,m}^{\text{self}}$ for NFEs based on the trainable matrices $W^Q_m \in \mathbb{R}^{d_h \times d_q}$ and $W^{K}_{m} \in \mathbb{R}^{d_h \times d_k}$ for head $m$ using Eq.~(\ref{eq:intqk}) as 
\begin{equation}
    \alpha_{i,j,m}^{\text{self}} = \frac{1}{\sqrt{d_k}} \left( h_i W^{Q}_{m} \right ) \left( h_j W^{K}_{m} \right ) ^T.
    \label{eq:intqk}
\end{equation}
Thereafter, the attention scores $\alpha^{\text{aux}}$ and $\alpha^{\text{self}}$ are fed into a three-layer MLP with structure ($2m \times 2m \times m$) to compute the synthesized multi-head attention scores as follows,
\begin{equation}
\resizebox{.91\linewidth}{!}{$
    \alpha_{i,j,1}^{\text{Synth}}, ..., \alpha_{i,j,{{m}}}^{\text{Synth}}\!=\!\text{MLP}\!\left( \alpha_{i,j,1}^{\text{self}}, ..., \alpha_{i,j,m}^{\text{self}}, \alpha_{i,j,1}^{\text{aux}}, ..., \alpha_{i,j,m}^{\text{aux}} \right).
    $}
\end{equation}
The scores are then normalized to $\tilde \alpha_{i,j, m}$ through Softmax, which are further used to calculate the attention values for each head as Eq.~(\ref{eq:value}). Finally, the outputs are given by Eq.~(\ref{eq:outputattn}) with trainable matrix $W^O \in \mathbb{R}^{md_v \times d_h}$ ($d_v = d_h / m$).
\begin{equation}
    \text{head}_{i,m} = \sum_{j=1}^{|V|}{\left. \tilde \alpha_{i,j, m} \left( h_j W^{V}_m \right)\right.},
    \label{eq:value}
\end{equation}
\begin{equation}
    {\tilde h}_i  =  \text{Concat} \left[ \text{head}_{i,1}, ..., \text{head}_{i,m} \right] W^O.
    \label{eq:outputattn}
\end{equation}
\paragraph{N2S Encoder.} We stack $L$ ($L$=3) encoders, each of which is the same as the Transformer encoder, except that the vanilla multi-head self-attention is replaced with our multi-head Synth-Att and we use the same instance normalization layer as \cite{ma2021learning}. Note that the auxiliary attention scores $\alpha_{i,j,m}^{\text{aux}}$ are only computed once and shared among all stacked encoders to reduce the computation costs.

\subsection{Decoder}
The N2S decoder first adopts the max-pooling layer in \cite{wu2021learning} to aggregate the global representation of all embeddings into each individual one as follows,
\begin{equation}
    {\hat h}_i = \tilde h_i^{(L)}W_h^{\textit{Local}} + \max\left[ \{\tilde h_i^{(L)}\}_{i=1}^{|V|} \right]W_h^{\textit{Global}}.
\end{equation}

\paragraph{Node-Pair Removal Decoder.}
Given the enhanced embeddings $\{\hat h_i\}_{i=1}^{|V|}$ and the set $\mathcal{H}(t,K)$, the \emph{removal} decoder outputs a categorical distribution over $n$ requests for removal action. In specific, it first computes a score $\lambda_i$ for each $x_i \in V$ indicating the closeness between node $x_i$ and its neighbors as
\begin{equation}
\resizebox{.89\linewidth}{!}{$
\begin{split}
        \lambda_{i} = ({\hat h}_{\text{pred}(x_i)} W^{Q}_\lambda)({\hat h}_i W^{K}_\lambda)^T + ({\hat h}_i W^{Q}_\lambda)({\hat h}_{\text{succ}(x_i)} W^{K}_\lambda)^T&\\
        - ({\hat h}_{\text{pred}(x_i)} W^{Q}_\lambda)({\hat h}_{\text{succ}(x_i)} W^{K}_\lambda)^T,&
        \end{split}
        $}
\end{equation}
where $\text{pred}(x_i)$ and $\text{succ}(x_i)$ refer to the predecessor and the successor nodes of $x_i$, respectively, and $W^{Q}_\lambda \in \mathbb{R}^{d_h \times d_h}$, $W^{K}_\lambda \in \mathbb{R}^{d_h \times d_h}$. We use multi-head technique to obtain $\lambda_{i,1}$ to $\lambda_{i,m}$. Then the decoder aggregates the scores for each pickup-delivery pair ($i^+,i^-$) based on a three-layer MLP$_\lambda$,
\begin{equation}
\resizebox{.89\linewidth}{!}{$
\begin{split}
    \tilde \Lambda_{(i^+,i^-)} = \text{MLP}_\lambda ( &\lambda_{i^+,1}, ...,\lambda_{i^+,m}, \lambda_{i^-, 1}, ..., \lambda_{i^-, m},\\
    & c(i), \mathds{1}_{\text{last}(1)=i}, \mathds{1}_{\text{last}(2)=i}, \mathds{1}_{\text{last}(3)=i} ),
    \end{split}
$}
\end{equation}
where the MLP structure is ($2m\!+\!4$, 32, 32, 1), scalar $c(i)$ counts the frequency of request $(i^+,i^-)$ being selected for removal in the past $K$ steps, and $\mathds{1}_{\text{last}(t)=i'}$ is a binary variable indicating whether request $i'$ was selected at the $t$-th last step. An activation layer ${\hat \Lambda}\!=\!C \cdot \text{Tanh}(\tilde \Lambda)$ is then applied ($C\!=\!6$), followed by Softmax to normalize the distribution which is then used to sample a node pair ($i^+,i^-$) as \emph{removal} action.

\paragraph{Node-Pair Reinsertion Decoder.}
Given a request $(i^+, i^-)$ for removal, the \emph{reinsertion} decoder outputs the joint distribution that reinserts the two nodes back to the solution. We first define two scores $\mu^\text{p}(x_\alpha, x_\beta)$ and $\mu^\text{s}(x_\alpha, x_\beta)$ for a node $x_\alpha$ indicating the degree of preference of accepting a node $x_\beta$ as its new predecessor and successor nodes, respectively,
\begin{equation}
\begin{split}
    \mu^\text{p}\left[x_\alpha, x_\beta \right] = ({\hat h}_{\alpha} W^{Q_\text{p}}_\mu)({\hat h}_{\beta} W^{K_\text{p}}_\mu)^T,\\
\mu^\text{s}\left[x_\alpha, x_\beta\right] = ({\hat h}_{\alpha} W^{Q_\text{s}}_\mu)({\hat h}_{\beta} W^{K_\text{s}}_\mu)^T,
\end{split}
\end{equation}
where $W^{Q_\text{p}}_\mu, W^{Q_\text{s}}_\mu\!\in\!\mathbb{R}^{d_h \times d_h}$, and $W^{K_\text{p}}_\mu, W^{K_\text{s}}_\mu\!\in\!\mathbb{R}^{d_h \times d_h}$. Again, we use multiple heads. Based on the scores, the decoder predicts the distribution of reinserting node $i^+$ after node $j$, and reinserting node $i^-$ after node $k$ using $\text{MLP}_\mu$,\!
\begin{equation}
\begin{split}
    \tilde \mu[j, k]=\text{MLP}_\mu\!(& \mu_1^\text{p}[i^+,{\text{succ}(j)}], ..., \mu_m^\text{p}[i^+,{\text{succ}(j)}],\\
    &\mu_1^\text{p}[i^-,{\text{succ}(k)}], ..., \mu_m^\text{p}[i^-,{\text{succ}(k)}],\\
    \mu_1^\text{s}[i^+,{j}], ...&, \mu_m^\text{s}[i^+,{j}], \mu_1^\text{s}[i^-,{k}], ..., \mu_m^\text{s}[i^-,{k}]),
\end{split}
\end{equation}
where the MLP structure is ($4m$, 32, 32, 1). Note that here $\text{pred}(\cdot)$ and $\text{succ}(\cdot)$ should be considered in the new solution where nodes $i^+, i^-$ have already been removed.  Afterwards, ${\hat \mu} = C \cdot \text{Tanh}(\tilde \mu)$ is applied and infeasible choices are masked as $-\infty$ before normalizing by Softmax. Finally, a node pair $(j,k)$, as the \emph{reinsertion} action, is sampled according to the resulting distribution to indicate the positions of reinserting the node-pair $(i^-,i^+)$ back to the solution.

\begin{algorithm}[t]
\caption{n-step PPO with CL strategy}
\label{ppo}
\textbf{Input}: policy $\pi_\theta$, critic $v_\phi$, PPO clipping threshold $\varepsilon$, learning rate $\eta_\theta$, $\eta_\phi$, learning rate decay $\beta$, epochs $E$, batches $B$, mini-batch $\kappa$, training steps $T_\text{train}$,  CL scalar $\rho^{CL}$

\begin{algorithmic}[1]
\FOR{$e = 1$ to $E$}
\FOR{$b = 1$ to $B$}
\STATE Generate training data $\mathcal{D}_b$ on the fly;
\STATE Initialize random solutions $\{\delta_{i}\}$ to $\mathcal{D}_b$;
\STATE Improve $\{\delta_{i}\}$ to $\{\delta'_{i}\}$ via $\pi_\theta$ for $T ={e}/{\rho^{CL}}$ steps;
\STATE Set initial state $s_0$ based on $\{\delta'_{i}\}$ and Eq.~(\ref{eq:states}); $t\!\gets\!0$;
\WHILE{$t < T_\text{train}$}
\STATE Get $\{(s_{t'},\!a_{t'},\!r_{t'})\}^{t + n}_{t' = t}$ where $a_{t'}\!\sim\!\pi_\theta(a_{t'}|s_{t'})$;
\STATE $t \gets t + n$, $\pi_{old} \gets \pi_\theta$, $v_{old} \gets v_\phi$;
\FOR{$k = 1$ to $\kappa$}
\STATE $\hat R_{t + 1} = v_\phi(s_{t + 1})$;
\FOR{$t' \in \{t, t - 1, ...,  t - n\}$}
\STATE $\hat R_{t'} \gets r_{t'} + \gamma \hat R_{t'+1}$;
\STATE $\hat A_{t'} \gets \hat R_{t'} - v_\phi(s_{t'})$;
\ENDFOR
\STATE Compute RL loss $J_{RL}(\theta)$ using Eq.~(\ref{eq:ppo}) and clipped critic loss $L_{BL}(\phi)$ using Eq.~(\ref{eq:BL});
\STATE $\theta \gets \theta + \eta_\theta \nabla J_{RL}(\theta)$; 
\STATE $\phi \gets \phi - \eta_\phi \nabla L_{BL}(\phi)$;
\ENDFOR
\ENDWHILE
\ENDFOR
\STATE $\eta_\theta \gets \beta \eta_\theta$, $\eta_\phi \gets \beta \eta_\phi$;
\ENDFOR
\end{algorithmic}
\end{algorithm}

\begin{algorithm}[t]
\caption{N2S-A Inference}
\label{alg:algorithma}
\textbf{Input}: Instance $\mathcal{I}$ with size $|V|$, policy $\pi_\theta$, maximum step $T$

\begin{algorithmic}[1] 
\FOR{$i=1, ..., \lfloor \frac{1}{2}|V| \rfloor$}
\STATE $\mathcal{I}_i \gets \mathcal{I}$;
\STATE $\mathcal{A}_i \gets \textbf{RandomShuffle}([\text{flip-x-y}, \text{1-x}, \text{1-y}, \text{rotate}])$;
\FOR{each augment method $j \in \mathcal{A}_i$}
\STATE $\varrho_j \gets$ $\textbf{RandomConfig}(j)$;
\STATE $\mathcal{I}_i \gets$ perform augment $j$ on $\mathcal{I}_i$ with config $\varrho_j$;
\ENDFOR
\ENDFOR
\STATE Solve all instances $\mathcal{I}_i$ in parallel with $\pi_\theta$ for $T$ steps;
\STATE \textbf{return} the best solution found among all $\mathcal{I}_i$
\end{algorithmic}
\end{algorithm}

\subsection{Training Algorithm}
As presented in Algorithm \ref{ppo}, the training algorithm for N2S is the same as \cite{ma2021learning}, namely the proximal policy optimization with a curriculum learning strategy. It learns a policy $\pi_\theta$ (our N2S) with the help of a critic  $v_\phi$ as follows.

\paragraph{Critic Network.} Given the embeddings $\{\tilde h_i^{(L)}\}_{i=0}^{|V|}$ from  policy network $\pi_\theta$, the critic network first enhances them by a vanilla multi-head attention layer (with $m=4$ heads) to get $\{ y_i \}_{i=0}^{|V|}$. The enhanced embeddings are then fed into a mean-pooling layer in \cite{wu2021learning} to aggregate the global representation of all embeddings into each individual one as
\begin{equation}
    {\hat y}_i =  y_iW_{v}^{\textit{Local}} + {\text{mean}}\left[ \{ y_i\}_{i=1}^{|V|} \right]W_{v}^{\textit{Global}},
    \label{eq:criticmean}
\end{equation}
where we use trainable matrices $W_{v}^{\textit{Local}},  W_{v}^{\textit{Global}} \in \mathbb{R}^{d_h\times\frac{d_h}{2}}$. Lastly, the state value $v(s_t)$ is output by a four-layer MLP in Eq.~(\ref{eq:valuemlp}) with structure (129, 128, 64, 1). Here, $f(\delta^*)$ is added as an additional input to $\text{MLP}_v$.
\begin{equation}
\resizebox{.89\linewidth}{!}{$
    v(s_t) = \text{MLP}_v \left({\text{max}}\!\left[ \{ \hat y_i\}_{i=1}^{|V|} \right], {\text{mean}}\!\left[ \{ \hat y_i\}_{i=1}^{|V|} \right], f(\delta^*)\right)
    $}
    \label{eq:valuemlp}
\end{equation}

We train N2S for $E$ epochs and $B$ batches per epoch, where the training dataset $\mathcal{D}_b$ is randomly generated on the fly with the uniform distribution (line 3). Following \cite{ma2021learning}, we introduce a similar CL strategy (in lines 4 to 6) to further ameliorate the training. In specific, it improves the randomly generated solutions $\{\delta_{i}\}$ to $\{\delta'_{i}\}$ by running the current policy $\pi_\theta$ for $T ={e}/{\rho^{CL}}$ steps. The improved solutions $\{\delta'_{i}\}$ with higher quality are then used to initialize the first state $s_0$. In such a way, the hardness level of neighborhood search is gradually increased per epoch following the idea of curriculum learning. The remaining algorithm follows the original design of the n-step PPO, where we present the reinforcement learning loss function in Eq.~(\ref{eq:ppo}), and the baseline loss function of critic in Eq.~(\ref{eq:BL}), respectively. Here, we clip the estimated value $\hat v(s_t)$ around the previous value estimates using $ v_\phi^{clip}(s_{t'})= clip\left[ v_\phi(s_{t'}), v_{old}(s_{t'})-\varepsilon, v_{old}(s_{t'})+\varepsilon \right]$.
\begin{equation}
\label{eq:ppo}
 \resizebox{.75\linewidth}{!}{$
 \begin{split}
    J_{RL}(\theta) = \frac{1}{n|\mathcal{D}_b|}\sum_{\mathcal{D}_b}\sum_{t' = t}^{t+n}min\left( \frac{\pi_\theta(a_{t'}|s_{t'})}{\pi_{old}(a_{t'}|s_{t'})}\hat A_{t'}, \right.\\
    \left. clip \left[ \frac{\pi_\theta(a_{t'}|s_{t'})}{\pi_{old}(a_{t'}|s_{t'})}, 1-\varepsilon, 1 + \varepsilon \right] \hat A_{t'}\right),
\end{split}   
    $}
\end{equation}
\begin{equation}
\label{eq:BL}
 \resizebox{.75\linewidth}{!}{$
 \begin{split}
    L_{BL}(\phi) = \frac{1}{n|\mathcal{D}_b|}\sum_{\mathcal{D}_b}\sum_{{t'} = t}^{{t}+n}max\left( \left| v_\phi(s_{t'}) - \hat R_{t'}\right|, \right. \\
    \left.\left| v_\phi^{clip}(s_{t'}) - \hat R_{t'}\right| \right)^2.
\end{split}  
$}
\end{equation}

\subsection{Diversity Enhancement}

To be more resistant to local minima, we further equip our N2S with an augmentation-based inference scheme, which leads to N2S-A in Algorithm \ref{alg:algorithma}. Such idea was originally explored in \cite{kwon2020pomo} for a neural \emph{construction} method, and we extend it to an \emph{improvement} one. The rationale is that an instance $\mathcal{I}$ can be transformed into different ones for searching while reserving the same optimal solution, e.g., rotating all locations of nodes by $\pi/2$ radian. For an instance of size $|V|$, our N2S-A performs $\lfloor\frac{1}{2}|V|\rfloor$ augments, each of which is generated by sequentially applying four preset invariant transformation operations with different orders and different configurations as listed in Table \ref{tab:transformation}. Note that although the mentioned transformations are conducted on instances defined in the Euclidean space, we believe that such an idea has favourable potential to be also exploited in non-Euclidean space, as long as there are proper invariant transformations for coordinates in the target space. Meanwhile, we use $K\!=\!|V|$ for training and $K'\!=\!\lfloor\frac{1}{2}|V|\rfloor$ for inference. This is because we found that when a specific $K$ in $\mathcal{H}(t, K)$ is used for training, a smaller $K$ during inference can improve the diversity of the solutions, thus better performance.

\begin{table}
\centering
\resizebox{0.48\textwidth}{!}{%
\begin{tabular}{@{}ccc@{}}
\toprule
Transformations &  Formulations & Configurations \\ \midrule
flip-x-y & $(x',y') = (y,x)$ & perform or skip  \\
1-x & $(x',y') = (1-x,y)$ & perform or skip  \\
1-y & $(x',y') = (x,1-y)$ & perform or skip  \\
rotate & 
$\begin{pmatrix}
x'\\ 
y'
\end{pmatrix} =
\begin{pmatrix}
x\cos\vartheta - y\sin\vartheta\\ 
x\sin\vartheta + y\cos\vartheta
\end{pmatrix}$ & $\vartheta \in \{0, \pi/2, \pi, 3\pi/2\}  $ \\ 
\bottomrule
\end{tabular}%
}
\caption{Descriptions of the four invariant transformations.}
\label{tab:transformation}
\end{table}

\section{Evaluation}
\label{sec:exp}

We design experiments to answer the following questions:
\begin{enumerate}
  \setlength{\itemsep}{1pt}
  \setlength{\parskip}{0pt}
  \setlength{\parsep}{0pt}
    \item How good is the proposed N2S approach against the baselines, including the state-of-the-art neural methods and the strong LKH3 solver? (see Table~\ref{tab:table1} and Table~\ref{tab:table2})
    \item Can Synth-Att reduce computation costs while achieving competitive performance to DAC-Att? (see Table \ref{tab:ab-1})
    \item How crucial are the proposed learnable node-pair removal and node-pair reinsertion decoders for achieving an efficient search? (see Table \ref{tab:ab-2})
    \item Can our N2S generalize well to benchmark instances that are different from training ones?  (see Table \ref{tab:generalization})
\end{enumerate}

\subsection{Setup}
We evaluate N2S on PDTSP and PDTSP-LIFO with three sizes $|V|=21,51,101$ following the conventions in \cite{kwon2020pomo,ma2021learning}, where the node coordinates of instances are randomly and uniformly generated in the unit square $[0, 1] \times [0, 1]$. For our N2S, the initial solution $\delta_0$ is sequentially constructed in a random fashion. Our experiments were conducted on a server equipped with 8 RTX 2080 Ti GPU cards and Intel E5-2680 CPU @ 2.4GHz. The training time of our N2S varies with problem sizes, \ie~around 1 day for $|V|\!=\!21$, 3 days for $|V|\!=\!51$, and 7 days for $|V|\!=\!101$, all of which are shorter than the baselines in Table~\ref{tab:baselines}.

\paragraph{Hyper-parameters.} Our N2S is trained with $E\!=\!200$ epochs and $B\!=\!20$ batches per epoch using batch size 600. We set $n\!=\!5$, $T_\text{train}\!=\!250$ for the $n$-step PPO with $\kappa\!=\!3$ mini-batch updates and a clip threshold $\epsilon\!=\!0.1$. Adam optimizer is used with learning rate $\eta_\theta\!=\!8\!\times\!10^{-5}$ for $\pi_\theta$, and  $\eta_\phi\!=\!2\!\times\!10^{-5}$ for $v_\phi$ (decayed $\beta\!=\!0.985$ per epoch). The discount factor $\gamma$ is set to 0.999 for both PDPs. We clip the gradient norm to be within 0.05, 0.15, 0.35, and set the curriculum learning $\rho^{CL}$ to 2, 1.5, 1 for the three problem sizes, respectively.

\subsection{Comparison Evaluation}
We compare our N2S with the state-of-the-art (SOTA) neural methods and the highly-optimized LKH3 solver. 

Regarding the former, we consider the SOTA \emph{improvement} method DACT~\cite{ma2021learning}\footnote{We use the \emph{insert} decoder which is the best for PDPs.} and the SOTA \emph{construction} method Heter-AM~\cite{li2021heterogeneous} (specially designed for PDPs). To make a fair comparison with our N2S-A (with the diversity enhancement), we upgrade Heter-AM to Heter-POMO, also given the known superiority of POMO to AM. In specific, we reserve the policy network in Heter-AM as the backbone while adopting the diverse rollouts and the data augmentation techniques in POMO~\cite{kwon2020pomo} to leverage the advantages of them for the best performance.
Each neural baseline is trained using the respective implementation code that is publicly available. For the upgraded Heter-POMO, we adapt and combine the model architecture from the original Heter-AM and the original POMO. The links to their original implementations, approximate training time for the size $|V|\!=\!101$ , and the used hyper-parameters are presented in Table \ref{tab:baselines}. For other hyper-parameters, we follow the recommendation in their papers. 
Regarding the latter, LKH3 is a strong heuristic (as reviewed in Section \ref{sec:relatedwork}) which is widely used as a baseline to benchmark neural methods in recent studies (e.g., \cite{ma2021learning,hottunglearning,li2021learning,wu2021learning}). We report its results with two settings of iterations, \ie~LKH (5k) and LKH (10k).

All baselines are evaluated on a test dataset with 2,000 instances, and we report the metrics of averaged objective values, averaged gaps to LKH3 (10K) and the total solving time. Note that it is hard to perform an absolutely fair time comparison between running Python codes on GPUs (neural methods) and running ANSI C codes on CPUs (LKH solver). Thus we follow the guidelines in \cite{accorsi2021guidelines} to perform the \emph{facilitate comparison} that lets each method make full use of the best settings on our machine. In particular, we report the time of LKH3 when running in parallel with 16 CPU cores and the time of each neural method when all 8 GPU cards are available (but do not need to be fully used).

\begin{table}
\centering
\resizebox{0.48\textwidth}{!}{%
\begin{tabular}{@{}rccc@{}}
\toprule
Method & Code  & \begin{tabular}[c]{@{}c@{}}Training Time\end{tabular} & Hyper-parameters\\
\midrule
Heter-AM & online\tablefootnote{\href{https://github.com/Demon0312/Heterogeneous-Attentions-PDP-DRL}{https://github.com/Demon0312/Heterogeneous-Attentions-PDP-DRL}}  & $\sim$ 20 days  &  \begin{tabular}[c]{@{}c@{}}train 800 epochs as per \\the original setting.\end{tabular} \\
\midrule
Heter-POMO  & online\tablefootnote{\href{https://github.com/yd-kwon/POMO}{https://github.com/yd-kwon/POMO}}  & $\sim$ 14 days   & \begin{tabular}[c]{@{}c@{}}train 2,000 epochs as per \\the original setting.\end{tabular}   \\
\midrule
DACT & online\tablefootnote{\href{https://github.com/yining043/VRP-DACT}{https://github.com/yining043/VRP-DACT}}  &  $\sim$ 10 days  & \begin{tabular}[c]{@{}c@{}} $\xi^{CL}=0.25,1,4 $ for sizes \\
$|v|$ = 21, 51, 101, respectively;\\$n\!=\!5$, $T_\text{train}\!=\!250$ (same as ours);\\train 200 epochs (same as ours).\end{tabular} \\
\bottomrule
\end{tabular}
}
\caption{Training details of the adopted neural baselines.}
\label{tab:baselines}
\end{table}

\begin{table*}
\centering
\resizebox{0.97\textwidth}{!}{%
\setlength{\tabcolsep}{1.8mm}
\begin{tabular}{@{}l|ccr|ccr|ccr@{}}
\toprule
\multirow{2}{*}{{{Methods}}} & \multicolumn{3}{c|}{{PDTSP-21}} & \multicolumn{3}{c|}{{PDTSP-51}} & \multicolumn{3}{c}{{PDTSP-101}} \\
 & {Obj. Value} & {Gap to LKH} &
 {Total Time} & {Obj. Value} & {Gap to LKH} & {Total Time} & {Obj. Value} & {Gap to LKH} & {Total Time} \\
 \midrule
 
LKH (5k) & 4.563 & 0.00\% & 3m & 6.866 & 0.06\% & 10m & 9.443 & 0.16\% & 49m \\
LKH (10k) & 4.563 & 0.00\% & 5m & 6.862 & 0.00\% & 19m & 9.428 & 0.00\% & 98m \\
\midrule
Heter-AM (gr.) & 4.655 & 2.02\% & (0s) & 7.333 & 6.86\% & (1s) & 10.348 & 9.76\% & (2s)\\
Heter-AM (5k) & 4.578 & 0.33\% & (33s) & 7.108 & 3.58\% & (1.5m) & 10.051 & 6.61\% & (5m)\\

DACT (1k) & 4.572 & 0.20\% & (18s) & 7.245 & 5.57\% & (29s) & 10.551 & 11.91\% & (51s) \\
DACT (2k) & 4.566 & 0.07\% & (37s) & 7.118 & 3.72\% & (1m) & 10.312 & 9.38\% & (1.5m) \\
DACT (3k) & 4.564 & 0.03\% & (1m) & 7.057 & 2.83\% & (1.5m) & 10.195 & 8.13\% & (2.5m) \\

N2S (1k) & 4.573 & 0.21\% & (21s) & 7.103 & 3.51\% & (31s) & 10.030 & 6.38\% & (1m) \\
N2S (2k) & 4.567 & 0.09\% & (42s) & 7.053 & 2.77\% & (1m) & 9.905 & 5.06\% & (2m) \\
N2S (3k) & 4.565 & 0.05\% & (1m) & 7.027 & 2.40\% & (1.5m) & 9.846 & 4.44\% & (3m) \\

\midrule

Heter-POMO (gr.) & {4.634} & 1.56\% & (0s) & 7.168 & 4.45\% & (1s) & 10.060 & 6.70\% & (2s) \\
Heter-POMO-A (gr.) & {4.584} & 0.46\% & (1s) & 6.995 & 1.93\% & (5s) & 9.681 & 2.68\% & (11s) \\
Heter-POMO-A (3k) & {4.564} & 0.03\% & (7m) & 6.916 & 0.77\% & (32m) & 9.567 & 1.47\% & (135m) \\

N2S-A (1k) & 4.563 & 0.01\% & (1m) & 6.865 & 0.03\% & (8m) & 9.475 & 0.50\% & (40m) \\
N2S-A (2k) & 4.563 & \textbf{0.00\%} & (2m) & 6.860 & \textbf{-0.03\%} & (16m) & 9.427 & \textbf{-0.01\%} & (80m) \\
N2S-A (3k) & 4.563 & \textbf{0.00\%} & (3m) & 6.860 & \textbf{-0.04\%} & (24m) & 9.409 & \textbf{-0.20\%} & (121m) \\ \bottomrule
\end{tabular}%
}

\caption{Comparison results for solving PDTSP instances of sizes $|V|=$ 21, 51, and 101.}
\label{tab:table1}
\end{table*}
\paragraph{Results on PDTSP.} 
Table \ref{tab:table1} shows the results on PDTSP. 
In the first group, we compare N2S with Heter-AM (\emph{greedy} and \emph{sampling}), and DACT. Compared to Heter-AM (5k), our N2S with only 1k steps attains lower gaps with less time for all sizes. Although DACT offers the best gap on PDTSP-21, its performance drops significantly as the problem size increases, partly because its decoder is less efficient than our node-pair \emph{removal} and \emph{reinsertion} ones when tackling larger-scale problems. Instead, our N2S achieves higher performance and consistently dominates DACT in terms of both the gaps and the time on PDTSP-51 and PDTSP-101.
In the second group, our augmented N2S-A is compared to the upgraded Heter-POMO method with three  variants\footnote{Heter-POMO (gr.), Heter-POMO-A (gr.),  Heter-POMO-A (3k) refer to: greedily~generate~$|V|$ solution; greedily generate $|V|$ solutions with 8 augments; sample $|V| \times$3k solutions with 8 augments.}.
It is shown that even with only 1k steps, our N2S-A attains a significantly smaller gap than all three Heter-POMO variants, by almost an order of magnitude. Although Heter-POMO-A (gr.) tends to be competitive with fast speed, the gap is hard to be further reduced by increasing inference time if we refer to Heter-POMO-A (3k). 
Moreover, our N2S-A (2k) keeps abreast of, or even slightly exceeds the strong LKH3 solver, achieving gaps of -0.03\% and -0.01\% with less time on PDTSP-51 and PDTSP-101, respectively. Those gaps are further reduced to -0.04\% and -0.20\% with more steps, \ie~3k.

\begin{table*}
\centering
\resizebox{0.97\textwidth}{!}{%
\setlength{\tabcolsep}{1.8mm}
\begin{tabular}{@{}l|ccr|ccr|ccr@{}}
\toprule
\multirow{2}{*}{{Methods}} & \multicolumn{3}{c|}{{PDTSP-LIFO-21}} & \multicolumn{3}{c|}{{PDTSP-LIFO-51}} & \multicolumn{3}{c}{{PDTSP-LIFO-101}} \\
 & {Obj. Value} & {Gap to LKH} &
 {Total Time} & {Obj. Value} & {Gap to LKH} & {Total Time} & {Obj. Value} & {Gap to LKH} & {Total Time} \\
 \midrule
LKH (5k) & 5.539 & 0.00\% & 1m & 10.218 & 0.17\% & 8m & 17.115 & 0.34\% & 33m \\
LKH (10k) & 5.539 & 0.00\% & 3m & 10.200 & 0.00\% & 16m & 17.057 & 0.00\% & 67m \\
\midrule

Heter-POMO (gr.) & 5.636 & 1.75\% & (0s) & 10.540 & 3.33\% & (1s) & 17.583 & 3.08\% & (2s)\\
Heter-POMO-A (gr.) & 5.567 & 0.51\% & (1s) & 10.353 & 1.50\% & (5s) & 17.276 & 1.29\% & (10s) \\
Heter-POMO-A (4k) & 5.545 & 0.12\% & (10m) & 10.209 & 0.09\% & (48m) & 16.890 & -0.98\% & (180m)\\


N2S-A (1k) & 5.539 & \textbf{0.00\%} & (3m) & 10.144 & \textbf{-0.55\%} & (13m) & 16.976 & -0.47\% & (54m) \\
N2S-A (2k) & 5.539 & \textbf{0.00\%} & (6m) & 10.137 & \textbf{-0.62\%} & (27m) & 16.859 & \textbf{-1.16\%} & (107m) \\
N2S-A (3k) & 5.539 & \textbf{0.00\%} & (9m) & 10.135 & \textbf{-0.64\%} & (40m) & 16.806 & \textbf{-1.47\%} & (161m)\\
\bottomrule
\end{tabular}%
}
\caption{Comparison results for solving PDTSP-LIFO instances of sizes $|V|=$ 21, 51, and 101.}
\label{tab:table2}
\end{table*}
\paragraph{Results on PDTSP-LIFO.} In Table \ref{tab:table2}, we report the results on PDTSP-LIFO. Due to the more constrained search space, DACT failed to work well (see Table \ref{tab:ab-2}). Therefore, we mainly compare our N2S-A with Heter-POMO (the best neural baseline in Table \ref{tab:table1}) and the LKH3 solver. As exhibited, the advantages of neural methods over LKH3 have been further enhanced on this harder problem, where our approach consistently outperforms Heter-POMO for all sizes. Compared to LKH3, our N2S-A with 3k steps presents superior performance again, and attains gaps of -0.64\% and -1.47\% on PDTSP-LIFO-51 and PDTSP-LIFO-101, respectively.


\subsection{Ablation Evaluation}
\paragraph{Effects of Different Encoding Methods.}
We replace the proposed Synth-Att in our N2S encoder with vanilla-Att and DAC-Att, respectively. Using only one GPU card, we report the number of model parameters, time, and gaps for solving 2,000 PDTSP-51 instances with 3k steps in Table \ref{tab:ab-1}. As exhibited, Synth-Att attains slightly smaller gaps than DAC-Att with much fewer computation costs, and considerably lower gaps than vanilla-Att with slight extra computation costs.

\begin{table}[]
\centering
\resizebox{0.43\textwidth}{!}{%
\begin{tabular}{@{}cccccc@{}}
\toprule
Att. in Encoders & Dim. & \# Param.(M) & Time(s) & Gap(\%)  \\ \midrule

Vanilla-Att & \multirow{3}{*}{64} & 0.18 (1.00{\small $\times$}) & 239 (1.00{\small $\times$}) & 4.66 \\
DAC-Att & & 0.31 (1.72{\small $\times$}) &  285 (1.19{\small $\times$}) & 2.89 \\
\textbf{Synth-Att} & & \textbf{0.19 (1.06{\small $\times$})} & \textbf{255 (1.07{\small $\times$})} &  \textbf{2.88}\\

\midrule
Vanilla-Att & \multirow{3}{*}{128} & 0.72 (1.00{\small $\times$}) & 322 (1.00{\small $\times$}) & 3.92 \\
DAC-Att & & 1.25 (1.73{\small $\times$})  & 400 (1.24{\small $\times$}) & 2.43 \\
\textbf{Synth-Att} & & \textbf{0.76 (1.06{\small $\times$})} & \textbf{340 (1.06{\small $\times$})} & \textbf{2.40} \\
\bottomrule
\end{tabular}%
}
\caption{Effects of different encoding methods.}
\label{tab:ab-1}
\end{table}

\paragraph{Effects of Different Decoding Methods.}
\begin{table}[t]
\centering
\resizebox{0.41\textwidth}{!}{%
\begin{tabular}{ccrr}
\toprule
\begin{tabular}[c]{@{}c@{}} Removal\\ Decoder\end{tabular} & \begin{tabular}[c]{@{}c@{}} Reinsertion\\ Decoder\end{tabular}& \begin{tabular}[c]{@{}r@{}}Gap(\%) on\\ PDTSP\end{tabular} & \begin{tabular}[c]{@{}r@{}}Gap(\%) on\\ PDTSP-LIFO\end{tabular} \\ \midrule
\xmark (random) & \xmark (random) & 210.82 & 112.37 \\
\xmark (random) & \xmark ($\epsilon$-greedy) & 18.03 & 17.87\\
\xmark ($\epsilon$-greedy) & \xmark (random) & 86.31 & 47.57 \\
\xmark ($\epsilon$-greedy) & \xmark ($\epsilon$-greedy) & 15.62 & 12.64 \\
\midrule
\cmark & \xmark (random) & 43.12 & 17.75\\
\cmark & \xmark ($\epsilon$-greedy) & 3.54 & 3.88\\
\xmark (random) & \cmark & 6.38 & 4.38 \\
\xmark ($\epsilon$-greedy) & \cmark & 8.42 & 6.28 \\
\midrule
\xmark (DACT) & \xmark (DACT) & 2.83 & 11.73 \\
\cmark & \cmark & \textbf{2.40} & \textbf{0.64} \\ \bottomrule
\end{tabular}%
}
\caption{Effects of different decoding methods.}
\label{tab:ab-2}
\end{table}

To highlight the desirability of our two decoders, we replace the trainable decoders with hand-crafted ones (\ie~random and the $\epsilon$-greedy with $\epsilon\!=\!0.1$) while ensuring feasibility. We gather the results on PDTSP-51 and PDTSP-LIFO-51 with 3k steps in Table~\ref{tab:ab-2} where mark `\cmark' means our proposed decoder is used (the one without augments) and mark `\xmark' means the decoder in the parentheses is used instead. As revealed, the methods of retaining at least one trainable decoder always attain much lower gaps than the ones equipped with only hand-crafted decoders. The method with both trainable decoders (\ie~N2S) achieves the best performance. We also notice that DACT fail to solve PDTSP-LIFO well. This might be because its decoder considers removing and reinserting only one node instead of a pickup-delivery node pair in each action, which is a serious limitation for more constrained PDPs. For example, on PDTSP-LIFO, over 92\% of the action space need to be masked for DACT, which leads to extremely low efficiency.

\begin{figure*}[htp]
{\hfill}
\begin{minipage}[t]{0.45\textwidth}
\centering
\subfloat[N201P3]{\includegraphics[width=2.35cm]{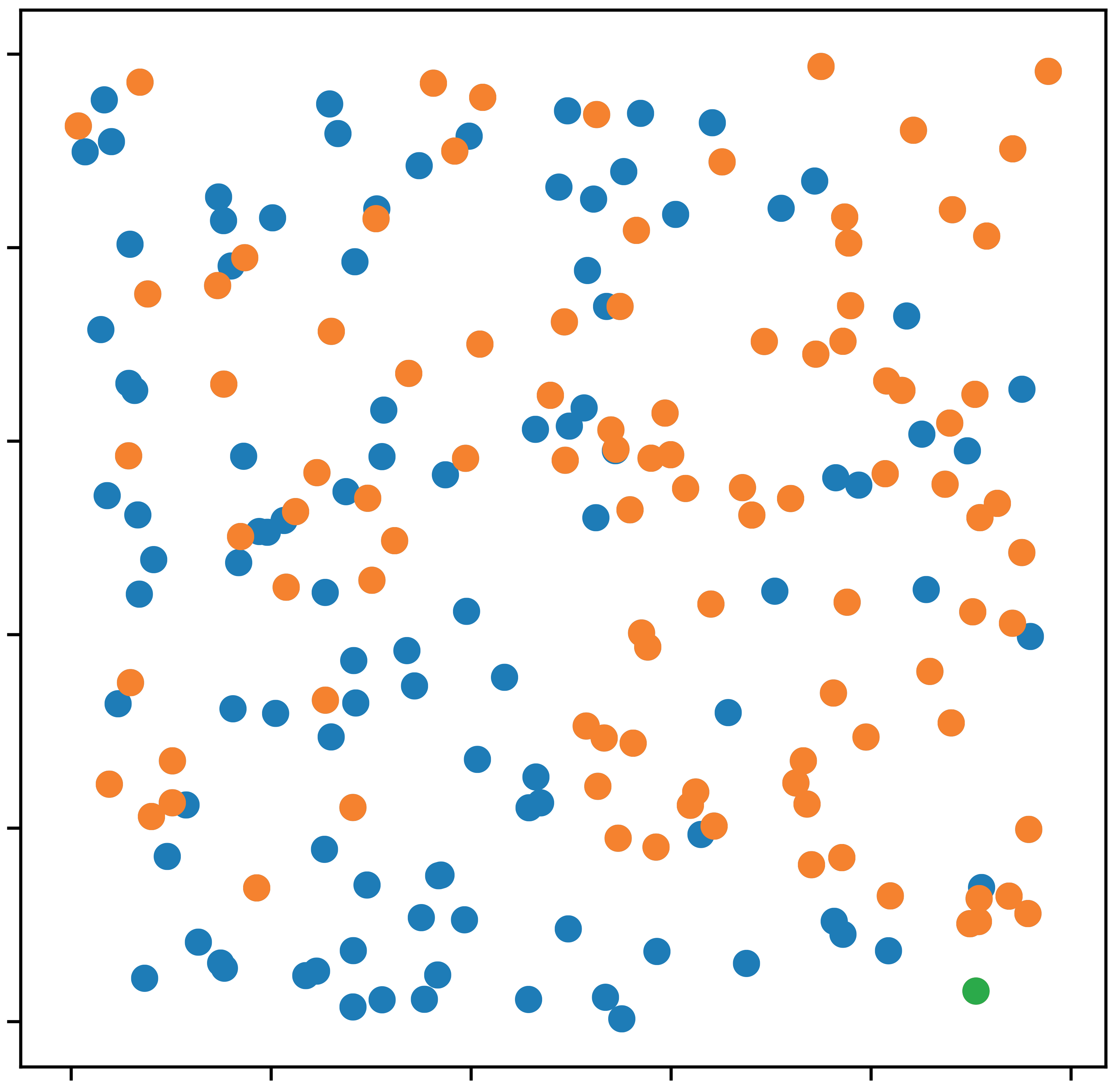}}
{\hfill}
\subfloat[N201P6]{\includegraphics[width=2.35cm]{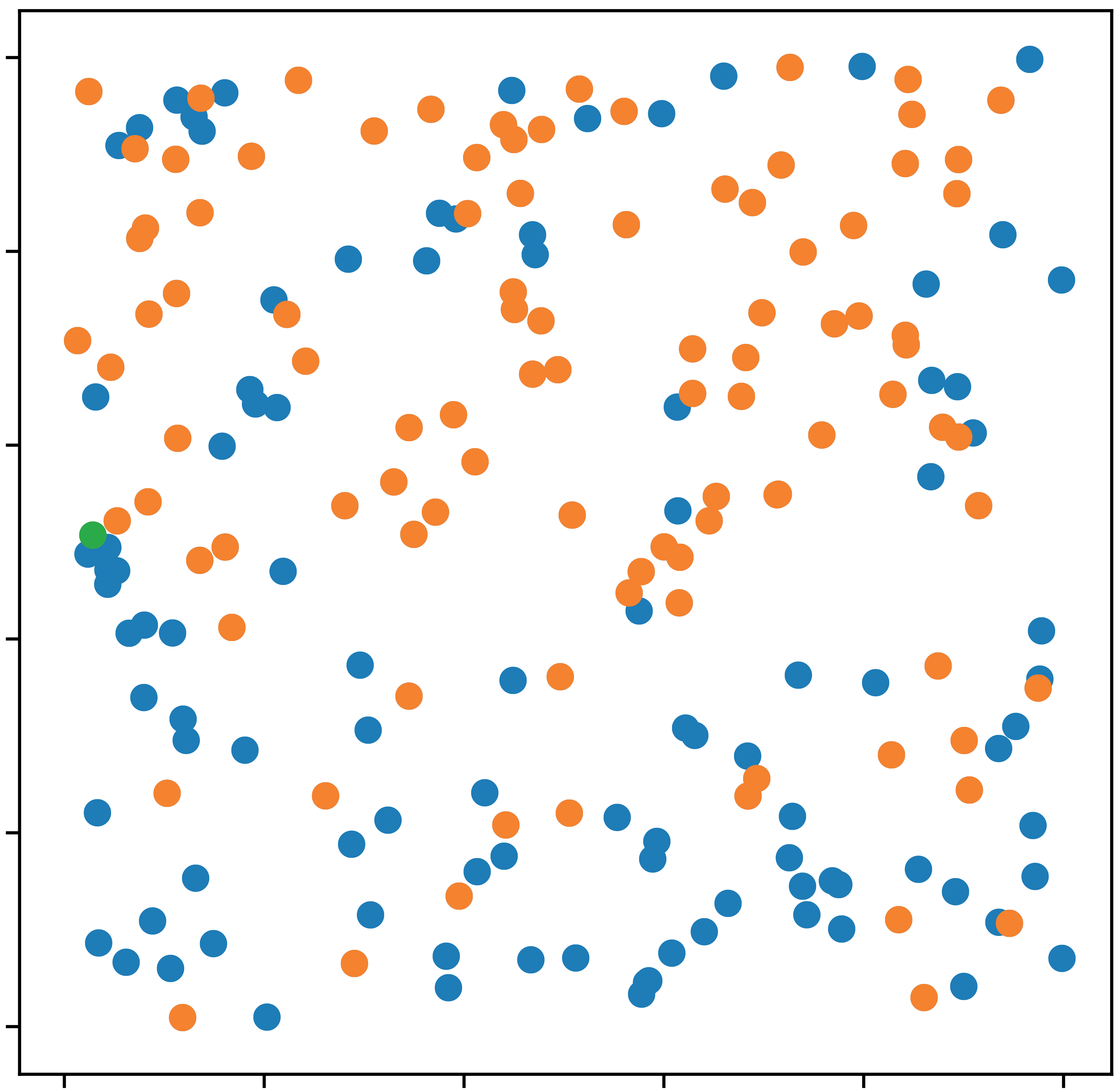}}
{\hfill}
\subfloat[N201P7]{\includegraphics[width=2.35cm]{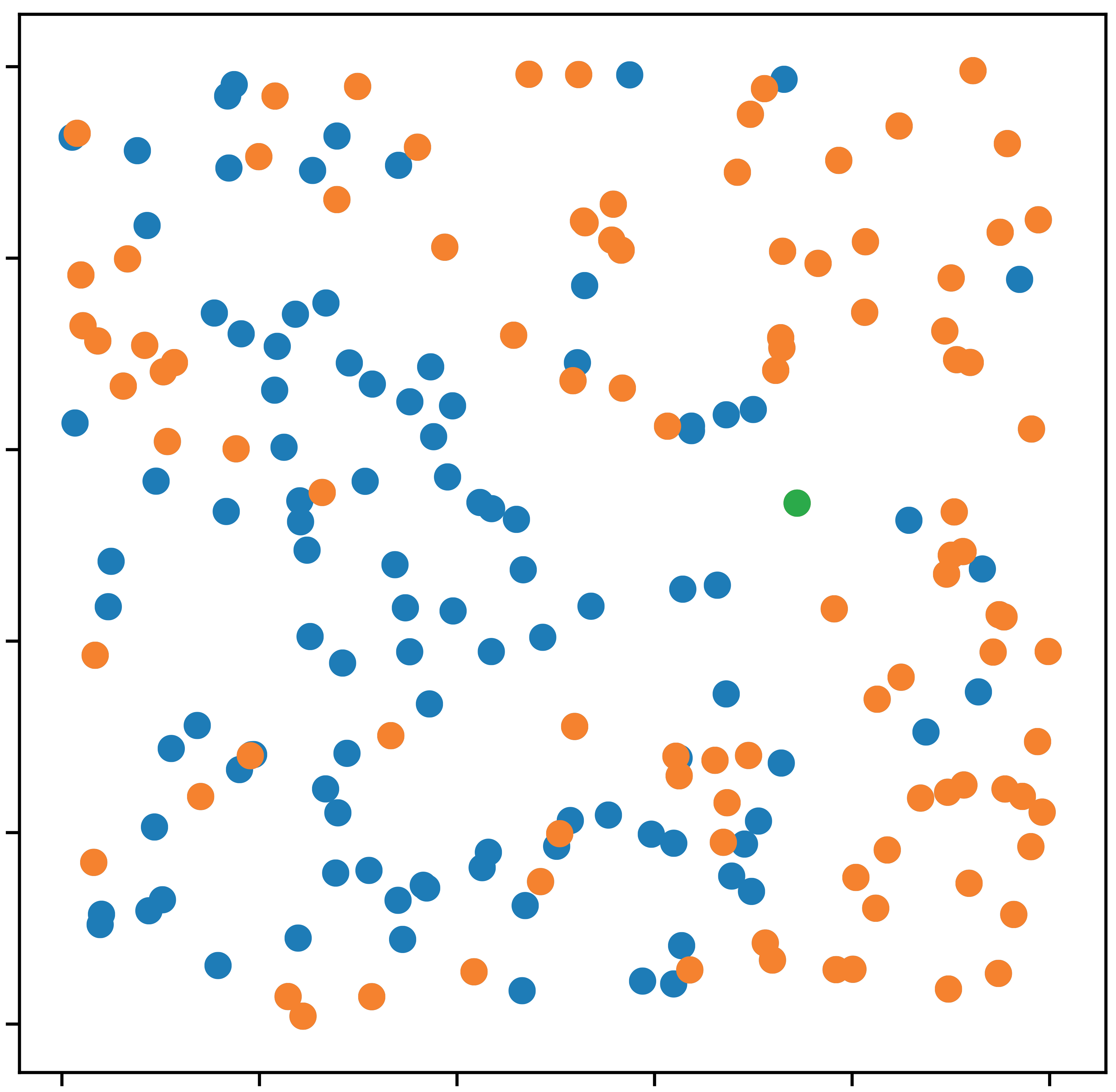}}
{\hfill}
\caption{Example instances from \protect\cite{renaud2002perturbation}.}
\label{fig:benchmarkpdtsp} 
\end{minipage}
{\hfill}
\begin{minipage}[t]{0.45\textwidth}
\centering
\subfloat[brd14051\_101]{\includegraphics[width=2.35cm]{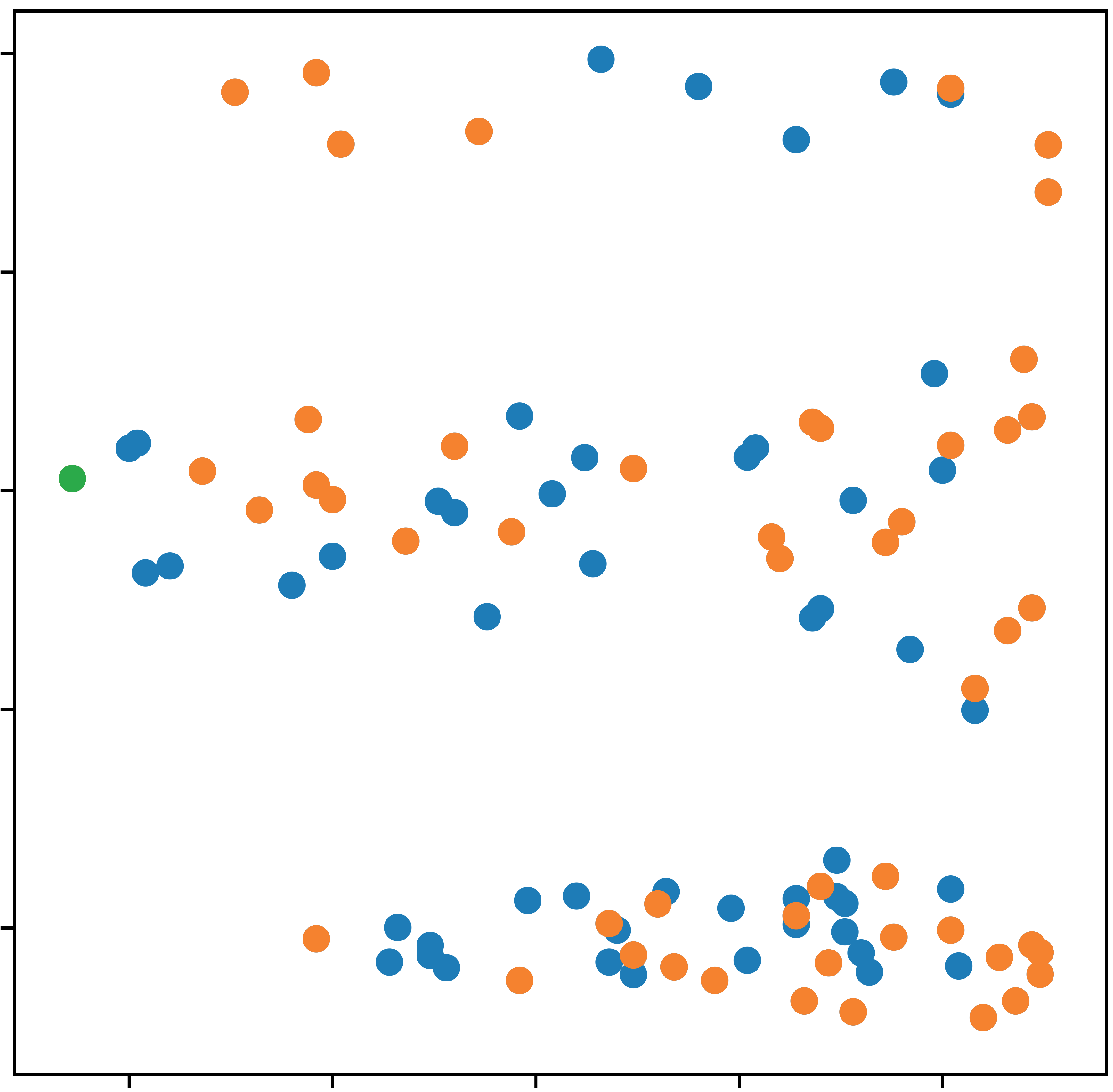}}
{\hfill}
\subfloat[pr1002\_101]{\includegraphics[width=2.35cm]{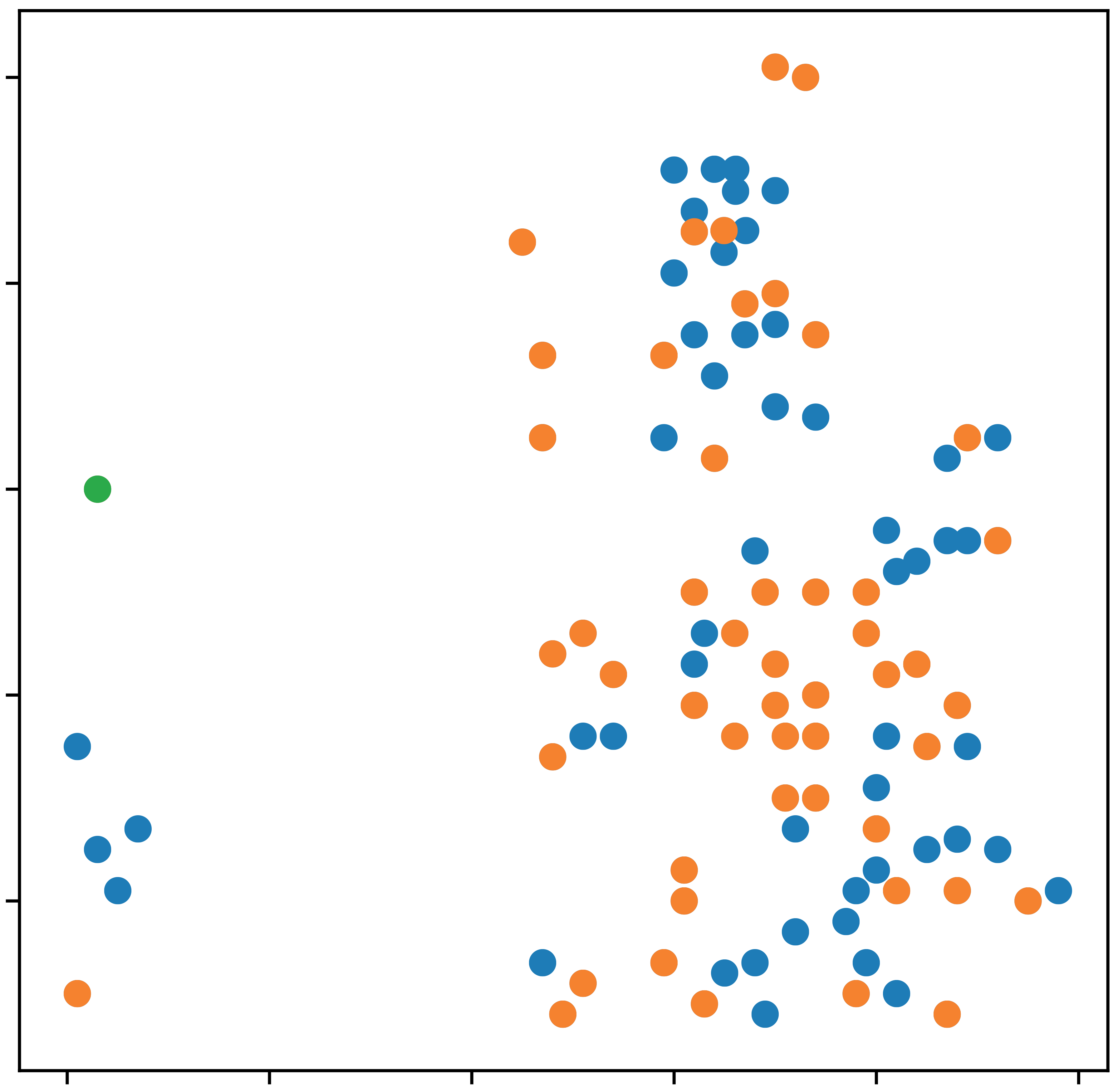}}
{\hfill}
\subfloat[fnl4461\_101]{\includegraphics[width=2.35cm]{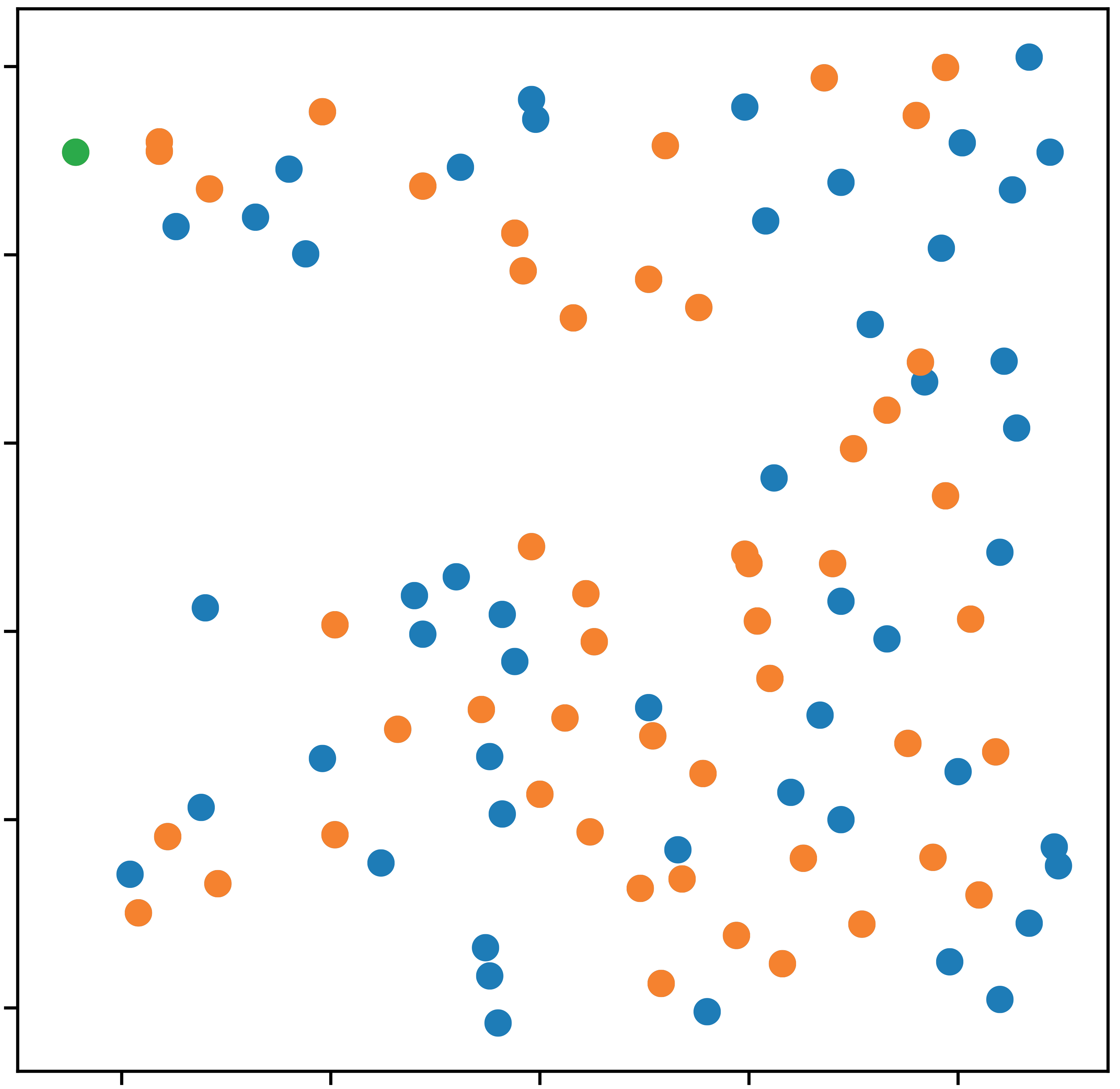}}
{\hfill}
\caption{Example instances from \protect\cite{carrabs2007variable}.}
\label{fig:benchmarkpdtsplifo} 
\end{minipage}
{\hfill}
\end{figure*}

\subsection{Generalization Evaluation}

We further evaluate our N2S on benchmark instances, including all the ones from \cite{renaud2002perturbation} for PDTSP and the ones with size $|V|\!\leq\!201$ from \cite{carrabs2007variable} for PDTSP-LIFO, which are largely different from our training ones, e.g., different sizes (i.e., 200 nodes) as shown in Figure \ref{fig:benchmarkpdtsp} and different node distributions as shown in Figure \ref{fig:benchmarkpdtsplifo}. In Table \ref{tab:generalization}, we report the best and the average gaps (with 10 independent runs) achieved by N2S-A and neural baseline Heter-POMO-A w.r.t. optimal solutions for PDTSP,  or heuristic baseline B1 \cite{carrabs2007variable} and B2 \cite{li2011tree} for PDTSP-LIFO. It can be seen that our N2S significantly outstrips Heter-POMO in all cases. Without re-training or leveraging other techniques, this is fairly hard to achieve because machine learning often suffers from a mediocre out-of-distribution zero-shot generalization \cite{zhou2021domain,li2021deep}.
The results also imply slight inferiority of our N2S to the LKH3 solver, given that LHK3 reports similar performance to the B2 baseline on those instances \cite{lkh3}.
Accordingly, we will focus on improving the out-of-distribution generalization for our N2S in future.

\begin{table}[t]
\centering
\resizebox{0.43\textwidth}{!}{%
\begin{tabular}{@{}ccccccc@{}}
\toprule
\multirow{2}{*}{Problem} &
  \multirow{2}{*}{$|V|$} &
  \multirow{2}{*}{Gaps to} &
  \multicolumn{2}{c}{Heter-POMO-A} &
  \multicolumn{2}{c}{N2S-A} \\ \cmidrule(lr){4-5}\cmidrule(lr){6-7}
 &    &   & Avg. & Best & Avg. & Best \\\midrule
\multirow{2}{*}{PDTSP} &
  101 &
  \multirow{2}{*}{Opt.} &
  1.64\% &
  1.46\% &
  \textbf{0.08\%} &
  \textbf{0.00\%} \\
 & 201 &    & 9.71\%  & 8.66\%    & \textbf{2.82\%}  & \textbf{2.19\%}    \\\midrule
\multirow{1}{*}{PDTSP} &
  \multirow{2}{*}{$\leq$101} &
  B1 &
  10.06\% &
  9.19\% &
 \textbf{0.85\%} &
  \textbf{-0.18\%} \\
  -LIFO
 &   & B2 & 11.46\% & 10.58\%   & \textbf{2.11\%}  & \textbf{1.06\%}   \\\bottomrule
\end{tabular}%
}
\caption{Generalization performance on benchmark instances.}
\label{tab:generalization}
\end{table}

\section{Conclusion}
\label{sec:conclusion}
We present an efficient N2S approach for PDPs. It utilizes a novel Synth-Att to synthesize node relationships from various types of solution features and exploits the node-pair removal and reinsertion decoders to tackle the precedence constraint. Extensive experiments on PDTSP and PDTSP-LIFO verified our design, where N2S achieves state-of-the-art performance among existing neural methods. Further equipped with a diversity enhancement scheme, it even becomes the first neural method to surpass LKH3 on synthesized PDP instances. In future, we will 1) deploy N2S as a low-level agent in the hierarchical framework of \cite{ma2021hierarchical} for dynamic PDPs, 2) combine N2S with a similar divide-and-conquer strategy in \cite{li2021learning} for much larger-scale instances, and 3) enhance the out-of-distribution generalization to exceed LKH3 on instances of arbitrary distributions.
\appendix
        
\section*{Acknowledgments}

This work was supported by the IEO Decentralised GAP project of Continuous Last-Mile Logistics (CLML) at SIMTech under Grant I22D1AG003. It was also supported in part by the National Natural Science Foundation of China under Grant 62102228, and in part by Guangdong Regional Joint Fund for Basic and Applied Research under Grant 2021B1515120078.
\bibliographystyle{named}
\bibliography{ijcai22}

\appendix

\onecolumn
\section{Full Results of Generalization}
We present more details for our generalization evaluation on benchmark datasets. 
For both Heter-POMO-A and N2S-A, the coordinates of instances are normalized to $[0,1]\times[0,1]$ as per training, and we adopted the  ``closest" model learned in Section \ref{sec:exp} to infer the instances, e.g., models trained on PDP-21 are used to infer PDP-25 instances, and models trained on PDP-101 are used to infer instances with $|V|\geq101$. We increase $C$ in our decoders during generalization since it boosts the performance.
Full results of Table \ref{tab:generalization} are gathered in Table \ref{tab:benchmark1} (PDTSP) and Table \ref{tab:benchmark2} (PDTSP-LIFO).
The gaps in Table \ref{tab:benchmark1} are computed w.r.t the optimal solutions and the gaps in Table \ref{tab:benchmark2} are computed w.r.t. heuristic baselines B1 \cite{carrabs2007variable} and B2 \cite{li2011tree}. In all cases, our N2S-A significantly outperforms Heter-POMO-A. In specific, we achieve the best gaps of 0.00\% ($|V|\!=\!101$), and 2.19\% ($|V|\!=\!201$) on PDTSP  while the gaps of Heter-POMO are 1.46\% ($|V|\!=\!101$) and 8.66\% ($|V|\!=\!201$); we achieve best gaps of -0.18\% (w.r.t.~B1) and 1.06\% (w.r.t.~B2) on PDTSP-LIFO while the gaps of Heter-POMO are 9.19\% (w.r.t.~B1) and 10.58\% (w.r.t.~B2). This further reveals the desirable generalization of our N2S approach.
\begin{table}[h]
\centering
\resizebox{0.85\textwidth}{!}{%
\setlength{\tabcolsep}{3mm}
\begin{tabular}{lcc|cccc|cccc}
\toprule \midrule
 &
   &
   &
  \multicolumn{2}{c}{Heter-POMO-A (3k)} &
  \multicolumn{2}{c|}{N2S-A (3k)} &
  \multicolumn{2}{c}{Heter-POMO-A (3k)} &
  \multicolumn{2}{c}{N2S-A (3k)} \\ \cmidrule(r){4-5}
  \cmidrule(r){6-7}
  \cmidrule(r){8-9}
  \cmidrule(r){10-11}
\multirow{-2}{*}{{Instances}} &
  \multirow{-2}{*}{{$|V|$}} &
  \multirow{-2}{*}{Optimal} &
  Avg. Cost &
  Best Cost &
  Avg. Cost &
  Best Cost &
  Avg. Gap(\%) &
  Best Gap(\%) &
  Avg. Gap(\%) &
  Best Gap(\%) \\
   \midrule\midrule
N101P1 &
  101 &
  799.0 &
  824.0 &
  820.0 &
  \textbf{800.0} &
  \textbf{799.0} &
  3.13{} &
  2.63{} &
  \textbf{0.13{}} &
  \textbf{0.00{}} \\
N101P2 &
  101 &
  729.0 &
  736.0 &
  735.0 &
  \textbf{730.0} &
  \textbf{729.0} &
  0.96{} &
  0.82{} &
  \textbf{0.14{}} &
  \textbf{0.00{}} \\
N101P3 &
  101 &
  748.0 &
  751.0 &
  751.0 &
  \textbf{748.0} &
  \textbf{748.0} &
  0.40{} &
  0.40{} &
  \textbf{0.00{}} &
  \textbf{0.00{}} \\
N101P4 &
  101 &
  807.0 &
  815.0 &
  814.0 &
  \textbf{808.0} &
  \textbf{807.0} &
  0.99{} &
  0.87{} &
  \textbf{0.12{}} &
  \textbf{0.00{}} \\
N101P5 &
  101 &
  783.0 &
  794.0 &
  791.0 &
  \textbf{783.0} &
  \textbf{783.0} &
  1.40{} &
  1.02{} &
  \textbf{0.00{}} &
  \textbf{0.00{}} \\
N101P6 &
  101 &
  755.0 &
  766.0 &
  763.0 &
  \textbf{755.0} &
  \textbf{755.0} &
  1.46{} &
  1.06{} &
  \textbf{0.00{}} &
  \textbf{0.00{}} \\
N101P7 &
  101 &
  767.0 &
  787.0 &
  787.0 &
  \textbf{768.0} &
  \textbf{767.0} &
  2.61{} &
  2.61{} &
  \textbf{0.13{}} &
  \textbf{0.00{}} \\
N101P8 &
  101 &
  762.0 &
  783.0 &
  782.0 &
  \textbf{764.0} &
  \textbf{762.0} &
  2.76{} &
  2.62{} &
  \textbf{0.26{}} &
  \textbf{0.00{}} \\
N101P9 &
  101 &
  766.0 &
  \textbf{766.0} &
  \textbf{766.0} &
  \textbf{766.0} &
  \textbf{766.0} &
  \textbf{0.00{}} &
  \textbf{0.00{}} &
  \textbf{0.00{}} &
  \textbf{0.00{}} \\
N101P10 &
  101 &
  754.0 &
  774.0 &
  773.0 &
  \textbf{754.0} &
  \textbf{754.0} &
  2.65{} &
  2.52{} &
  \textbf{0.00{}} &
  \textbf{0.00{}} \\
   \midrule
 \textbf{Average} & \multicolumn{1}{r}{}
   &
  {767.0*} &
  779.6 &
  778.2 &
  \textbf{767.6} &
  \textbf{767.0*} &
  1.64{} &
  1.46{} &
  \textbf{0.08{}} &
  \textbf{0.00{}} \\
   \midrule \midrule
N201P1 &
  201 &
  1039.0 &
  1109.0 &
  1095.0 &
  \textbf{1068.0} &
  \textbf{1059.0} &
  6.74{} &
  5.39{} &
  \textbf{2.79{}} &
  \textbf{1.92{}} \\
N201P2 &
  201 &
  1086.0 &
  1117.0 &
  1114.0 &
  \textbf{1102.0} &
  \textbf{1099.0} &
  2.85{} &
  2.58{} &
  \textbf{1.47{}} &
  \textbf{1.20{}} \\
N201P3 &
  201 &
  1070.0 &
  1198.0 &
  1177.0 &
  \textbf{1112.0} &
  \textbf{1107.0} &
  11.96{} &
  10.00{} &
  \textbf{3.93{}} &
  \textbf{3.46{}} \\
N201P4 &
  201 &
  1050.0 &
  1114.0 &
  1108.0 &
  \textbf{1073.0} &
  \textbf{1066.0} &
  6.10{} &
  5.52{} &
  \textbf{2.19{}} &
  \textbf{1.52{}} \\
N201P5 &
  201 &
  1052.0 &
  1187.0 &
  1169.0 &
  \textbf{1094.0} &
  \textbf{1078.0} &
  12.83{} &
  11.12{} &
  \textbf{3.99{}} &
  \textbf{2.47{}} \\
N201P6 &
  201 &
  1059.0 &
  1120.0 &
  1111.0 &
  \textbf{1077.0} &
  \textbf{1072.0} &
  5.76{} &
  4.91{} &
  \textbf{1.70{}} &
  \textbf{1.23{}} \\
N201P7 &
  201 &
  1037.0 &
  1170.0 &
  1164.0 &
  \textbf{1076.0} &
  \textbf{1065.0} &
  12.83{} &
  12.25{} &
  \textbf{3.76{}} &
  \textbf{2.70{}} \\
N201P8 &
  201 &
  1079.0 &
  1215.0 &
  1196.0 &
  \textbf{1112.0} &
  \textbf{1108.0} &
  12.60{} &
  10.84{} &
  \textbf{3.06{}} &
  \textbf{2.69{}} \\
N201P9 &
  201 &
  1050.0 &
  1208.0 &
  1198.0 &
  \textbf{1076.0} &
  \textbf{1073.0} &
  15.05{} &
  14.10{} &
  \textbf{2.48{}} &
  \textbf{2.19{}} \\
N201P10 &
  201 &
  1085.0 &
  1198.0 &
  1192.0 &
  \textbf{1116.0} &
  \textbf{1112.0} &
  10.41{} &
  9.86{} &
  \textbf{2.86{}} &
  \textbf{2.49{}} \\
 \midrule
\textbf{Average} &
   \multicolumn{1}{r}{} &
  {1060.7*} &
  1163.6 &
  1152.4 &
  \textbf{1090.6} &
  \textbf{1083.9} &
  9.71{} &
  8.66{} &
  \textbf{2.82{}} &
  \textbf{2.19{}} \\  \midrule\bottomrule
\end{tabular}%
}
\caption{Generalization performance on benchmark instances from \protect\cite{renaud2002perturbation} for PDTSP using the trained model in Section \ref{sec:exp}.}
\label{tab:benchmark1}
\end{table}

\begin{table}[h]
\centering
\resizebox{0.95\textwidth}{!}{%
\begin{tabular}{@{}rrrr|rrrr|rrrr|rrrr@{}}
\toprule
\midrule
\multirow{3}{*}{Instances} &
  \multicolumn{1}{c}{\multirow{3}{*}{$|V|$}} &
  \multirow{3}{*}{B1(2007)} &
   \multirow{3}{*}{B2(2011)} &
  \multicolumn{2}{c}{Heter-POMO-A (3k)} &
  \multicolumn{2}{c|}{N2S-A (3k)} &
  \multicolumn{2}{c}{Heter-POMO-A (3k)} &
  \multicolumn{2}{c|}{N2S-A (3k)} &
  \multicolumn{2}{c}{Heter-POMO-A (3k)} &
  \multicolumn{2}{c}{N2S-A (3k)} \\ \cmidrule(l){5-6} \cmidrule(l){7-8} \cmidrule(l){9-10} \cmidrule(l){11-12}   \cmidrule(l){13-14} \cmidrule(l){15-16}  
\multicolumn{1}{r}{} &
  \multicolumn{1}{c}{} &
  \multicolumn{1}{c}{} &
  \multicolumn{1}{c|}{} &
  \multicolumn{1}{c}{Avg. Cost} &
  \multicolumn{1}{c}{Best Cost} &
  \multicolumn{1}{c}{Avg. Cost} &
  \multicolumn{1}{c|}{Best Cost} &
  \multicolumn{1}{c}{\begin{tabular}[c]{@{}c@{}}Avg. Gap\\  to B1(\%)\end{tabular}} &
  \multicolumn{1}{c}{\begin{tabular}[c]{@{}c@{}}Best Gap\\  to B1(\%)\end{tabular}} &
  \multicolumn{1}{c}{\begin{tabular}[c]{@{}c@{}}Avg. Gap\\  to B1(\%)\end{tabular}} &
  \multicolumn{1}{c|}{\begin{tabular}[c]{@{}c@{}}Best Gap\\  to B1(\%)\end{tabular}} &
  \multicolumn{1}{c}{\begin{tabular}[c]{@{}c@{}}Avg. Gap\\  to B2(\%)\end{tabular}} &
  \multicolumn{1}{c}{\begin{tabular}[c]{@{}c@{}}Best Gap\\  to B2(\%)\end{tabular}} &
  \multicolumn{1}{c}{\begin{tabular}[c]{@{}c@{}}Avg. Gap\\  to B2(\%)\end{tabular}} &
  \multicolumn{1}{c}{\begin{tabular}[c]{@{}c@{}}Best Gap\\  to B2(\%)\end{tabular}}\\
 \midrule\midrule
brd14051 &
  25 &
  4682.2 &
  4672.0 &
  4705.0 &
  4705.0 &
  \textbf{4680.0} &
  \textbf{4672.0} &
  0.49 &
  0.49 &
  \textbf{-0.05} &
  \textbf{-0.22} &
  0.71 &
  0.71 &
  \textbf{0.17} &
  \textbf{0.00} \\
 &
  51 &
  7763.2 &
  7740.0 &
  8276.0 &
  8201.0 &
  \textbf{7948.0} &
  \textbf{7828.0} &
  6.61 &
  5.64 &
  \textbf{2.38} &
  \textbf{0.83} &
  6.93 &
  5.96 &
  \textbf{2.69} &
  \textbf{1.14} \\
 &
  75 &
  7309.1 &
  7232.4 &
  9059.0 &
  8938.0 &
  \textbf{7775.0} &
  \textbf{7554.0} &
  23.94 &
  22.29 &
  \textbf{6.37} &
  \textbf{3.35} &
  25.26 &
  23.58 &
  \textbf{7.50} &
  \textbf{4.45} \\
 &
  101 &
  10005.2 &
  9735.0 &
  13315.0 &
  13000.0 &
  \textbf{10539.0} &
  \textbf{10370.0} &
  33.08 &
  29.93 &
  \textbf{5.34} &
  \textbf{3.65} &
  36.77 &
  33.54 &
  \textbf{8.26} &
  \textbf{6.52} \vspace{0.3cm}\\
pr1002 &
  25 &
  16221.0 &
  16221.0 &
  \textbf{16221.0} &
  \textbf{16221.0} &
  \textbf{16221.0} &
  \textbf{16221.0} &
  \textbf{0.00} &
  \textbf{0.00} &
  \textbf{0.00} &
  \textbf{0.00} &
  \textbf{0.00} &
  \textbf{0.00} &
  \textbf{0.00} &
  \textbf{0.00} \\
 &
  51 &
  31186.7 &
  30936.0 &
  30936.0 &
  30936.0 &
  \textbf{30936.0} &
  \textbf{30936.0} &
  -0.80 &
  -0.80 &
  \textbf{-0.80} &
  \textbf{-0.80} &
  0.00 &
  0.00 &
  \textbf{0.00} &
  \textbf{0.00} \\
 &
  75 &
  46911.0 &
  46673.0 &
  47404.0 &
  47202.0 &
  \textbf{47284} &
  \textbf{46923} &
  1.05 &
  0.62 &
  \textbf{0.80} &
  \textbf{0.03} &
  1.57 &
  1.13 &
  \textbf{1.31} &
  \textbf{0.54} \\
 &
  101 &
  63611.1 &
  61433.0 &
  \textbf{62569.0} &
  62565.0 &
  62787.0 &
  \textbf{62353.0} &
  \textbf{-1.64} &
  -1.64 &
  -1.30 &
  \textbf{-1.98} &
  \textbf{1.85} &
  1.84 &
  2.20 &
  \textbf{1.50} \vspace{0.3cm}\\
fnl4461 &
  25 &
  2168.0 &
  2168.0 &
  \textbf{2168.0} &
  \textbf{2168.0} &
  \textbf{2168.0} &
  \textbf{2168.0} &
  \textbf{0.00} &
  \textbf{0.00} &
  \textbf{0.00} &
  \textbf{0.00} &
  \textbf{0.00} &
  \textbf{0.00} &
  \textbf{0.00} &
  \textbf{0.00} \\
 &
  51 &
  4020.0 &
  4020.0 &
  4038.0 &
  4037.0 &
  \textbf{4020.0} &
  \textbf{4020.0} &
  0.45 &
  0.42 &
  \textbf{0.00} &
  \textbf{0.00} &
  0.45 &
  0.42 &
  \textbf{0.00} &
  \textbf{0.00} \\
 &
  75 &
  5865.0 &
  5739.0 &
  6118.0 &
  6038.0 &
  \textbf{5905.0} &
  \textbf{5763.0} &
  4.31 &
  2.95 &
  \textbf{0.68} &
  \textbf{-1.74} &
  6.60 &
  5.21 &
  \textbf{2.89} &
  \textbf{0.42} \\
 &
  101 &
  8852.8 &
  8562.0 &
  9186.0 &
  9065.0 &
  \textbf{8827.0} &
  \textbf{8702.0} &
  3.76 &
  2.40 &
  \textbf{-0.29} &
  \textbf{-1.70} &
  7.29 &
  5.87 &
  \textbf{3.10} &
  \textbf{1.64} \vspace{0.3cm}\\
d18512 &
  25 &
  4683.4 &
  4672.0 &
  4707.0 &
  4705.0 &
  \textbf{4680.0} &
  \textbf{4672.0} &
  0.50 &
  0.46 &
  \textbf{-0.07} &
  \textbf{-0.24} &
  0.75 &
  0.71 &
  \textbf{0.17} &
  \textbf{0.00} \\
 &
  51 &
  7565.6 &
  7502.0 &
  8215.0 &
  8126.0 &
  \textbf{7696.0} &
  \textbf{7519.0} &
  8.58 &
  7.41 &
  \textbf{1.72} &
  \textbf{-0.62} &
  9.50 &
  8.32 &
  \textbf{2.59} &
  \textbf{0.23} \\
 &
  75 &
  8781.5 &
  8629.0 &
  10282.0 &
  10215.0 &
  \textbf{8884.0} &
  \textbf{8802.0} &
  17.09 &
  16.32 &
  \textbf{1.17} &
  \textbf{0.23} &
  19.16 &
  18.38 &
  \textbf{2.96} &
  \textbf{2.00} \\
 &
  101 &
  10332.4 &
  10256.4 &
  13499.0 &
  13218.0 &
  \textbf{10729.0} &
  \textbf{10555.0} &
  30.65 &
  27.93 &
  \textbf{3.84} &
  \textbf{2.15} &
  31.62 &
  28.88 &
  \textbf{4.61} &
  \textbf{2.91} \vspace{0.3cm}\\
d15112 &
  25 &
  93981.0 &
  93981.0 &
  \textbf{93981.0} &
  \textbf{93981.0} &
  \textbf{93981.0} &
  \textbf{93981.0} &
  \textbf{0.00} &
  \textbf{0.00} &
  \textbf{0.00} &
  \textbf{0.00} &
  \textbf{0.00} &
  \textbf{0.00} &
  \textbf{0.00} &
  \textbf{0.00} \\
 &
  51 &
  143575.2 &
  142113.0 &
  144120.0 &
  143716.0 &
  \textbf{142113.0} &
  \textbf{142113.0} &
  0.38 &
  0.10 &
  \textbf{-1.02} &
  \textbf{-1.02} &
  1.41 &
  1.13 &
  \textbf{0.00} &
  \textbf{0.00} \\
 &
  75 &
  201385.4 &
  199047.8 &
  206442.0 &
  204944.0 &
  \textbf{200004.0} &
  \textbf{199076.0} &
  2.51 &
  1.77 &
  \textbf{-0.69} &
  \textbf{-1.15} &
  3.71 &
  2.96 &
  \textbf{0.48} &
  \textbf{0.01} \\
 &
  101 &
  276876.8 &
  266925.3 &
  272784.0 &
  273693.0 &
  \textbf{269160.0} &
  \textbf{267305.0} &
  -1.48 &
  -1.15 &
  \textbf{-2.79} &
  \textbf{-3.46} &
  2.19 &
  2.54 &
  \textbf{0.84} &
  \textbf{0.14} \vspace{0.3cm}\\
nrw1379 &
  25 &
  3194.8 &
  3192.0 &
  \textbf{3192.0} &
  \textbf{3192.0} &
  \textbf{3192.0} &
  \textbf{3192.0} &
  \textbf{-0.09} &
  \textbf{-0.09} &
  \textbf{-0.09} &
  \textbf{-0.09} &
  \textbf{0.00} &
  \textbf{0.00} &
  \textbf{0.00} &
  \textbf{0.00} \\
 &
  51 &
  5095.0 &
  5055.0 &
  6369.0 &
  6517.0 &
  \textbf{5086.0} &
  \textbf{5056.0} &
  25.00 &
  27.91 &
  \textbf{-0.18} &
  \textbf{-0.77} &
  25.99 &
  28.92 &
  \textbf{0.61} &
  \textbf{0.02} \\
 &
  75 &
  6865.1 &
  6831.0 &
  9647.0 &
  9272.0 &
  \textbf{7081.0} &
  \textbf{6960.0} &
  40.52 &
  35.06 &
  \textbf{3.14} &
  \textbf{1.38} &
  41.22 &
  35.73 &
  \textbf{3.66} &
  \textbf{1.89} \\
 &
  101 &
  10197.5 &
  9889.4 &
  13898.0 &
  13592.0 &
  \textbf{10330.0} &
  \textbf{9996.0} &
  36.29 &
  33.29 &
  \textbf{1.30} &
  \textbf{-1.98} &
  40.53 &
  37.44 &
  \textbf{4.46} &
  \textbf{1.08} \\
\midrule
\multicolumn{1}{c}{\multirow{1}{*}{\textbf{Average}}} &
  \multicolumn{1}{r}{} &
  42518.9 &
  41740.6 &
  43388.7 &
  43263.3 &
  \textbf{42123.2} &
  \textbf{41893.3} &
  10.06 &
  9.19 &
  \textbf{0.85} &
  \textbf{-0.18} &
  11.46 &
  10.58 &
  \textbf{2.11} &
  \textbf{1.06} \\
\midrule\bottomrule
\end{tabular}%
}
\caption{Generalization performance on benchmark instances from \protect\cite{li2011tree} for PDTSP-LIFO using the trained model in Section 
\ref{sec:exp}.}
\label{tab:benchmark2}
\end{table}

\section{Dealing with Capacity Constraint}
Our N2S is generic to the capacity constraint, similar to DACT for handling capacity in CVRP \cite{ma2021learning}. Specifically, we can, 1) make copies of depots (i.e., dummy depots) so that N2S can search solutions with different numbers of vehicles automatically; 2) add capacity/demand features to NFEs; 3) mask out infeasible choices in the Reinsertion decoder; and 4) use diversity enhancement as usual since it only affects the node coordinates. Nevertheless, the capacity in PDP might not be so crucial as the vehicle may always alternatively load or unload the goods.

\end{document}